\documentclass[runningheads]{llncs}
\usepackage[T1]{fontenc}
\usepackage{graphicx}
\usepackage{subfig}
\usepackage{cite}
\usepackage{amsmath,amssymb,amsfonts}
\usepackage{algorithmic}
\usepackage{textcomp}
\usepackage{verbatim}
\usepackage{multirow}
\usepackage{xcolor}
\usepackage[ruled,noline,nofillcomment]{algorithm2e}
\SetKwIF{If}{ElseIf}{Else}{if}{:}{elif}{else:}{}
\begin{document}
\title{Game and Reference: Policy Combination Synthesis for Epidemic Prevention and Control}
%
%
\author{Zhiyi Tan\inst{1} \and
Bingkun Bao\inst{1} }
\authorrunning{F. Author et al.}
%
\institute{Nanjing University of Post and Telecommunication, China\\ 
\email{tzy, bingkunbao@njupt.edu.cn}}

\maketitle              
\begin{abstract}
In recent years, epidemic policy-making models are increasingly being used to provide reference for governors on prevention and control policies against catastrophic epidemics such as SARS, H1N1 and COVID-19. Existing studies are currently constrained by two issues: First, previous methods develop policies based on effect evaluation, since few of factors in real-world decision-making can be modeled, the output policies will then easily become extreme. Second, the subjectivity and cognitive limitation of human make the historical policies not always optimal for the training of decision models. To these ends, we present a novel \textbf{P}olicy \textbf{C}ombination \textbf{S}ynthesis (PCS) model for epidemic policy-making. Specially, to prevent extreme decisions, we introduce adversarial learning between the model-made policies and the real policies to force the output policies to be more human-liked. On the other hand, to minimize the impact of sub-optimal historical policies, we employ contrastive learning to let the model draw on experience from the best historical policies under similar scenarios. Both adversarial and contrastive learning are adaptive based on the comprehensive effects of real policies to ensure the model always learns useful information.
Extensive experiments on real-world data prove the effectiveness of the proposed model.

\end{abstract}
\section{Introduction}
It has been witnessed that the global infectious disease has become one of the biggest threats to human society. Whenever an epidemic occurs, the prevention and control policies including the well-known \textbf{N}on-\textbf{P}harmaceutical \textbf{I}nterventions (NPIs) play a crucial role in epidemic containment. A proper epidemic prevention and control policy should achieve a balance between the containment effect and its impact on society and economy. Developing such a policy requires a comprehensive analysis of multiple aspects of large-scale data such as epidemics, social activities, economics et al, which is almost impossible for human decision-makers. Thus, models for epidemic policy-making have been extensively proposed to provide supporting suggestions.

Existing models on epidemic policy-making are mostly derived from epidemic prediction models. 
Commonly applied epidemic prediction models in previous studies include the classic epidemic model such as SEIR and SEAIR \cite{ANNAS2020110072}\cite{10001858}, traditional sequential prediction models such as MLR and ARIMA \cite{info11090454}\cite{HERNANDEZMATAMOROS2020106610}, and the neural networks for sequential modeling such as LSTM \cite{KARA2021115153}. Though achieved initial success, existing models are still not yet competent as an assistant in government epidemic policy-making because of the following reasons: First, policies made by the existing models are usually headed for the best epidemic containment effect evaluation. However, in real-world policy-making, there are complex factors that need to be considered while few of them can be directly modeled by models (social stability, living support, operation of functional departments et al). In this case, models will be prompt to simply increase the policies' intensities for better evaluation scores, thereby harming social and economic development. Second, manually determined policies are constrained by the knowledge and experience of human decision-makers. Thus, when learning to develop policies for a given region, the region's historical policies may be far from optimal references for the model and may mislead the model's training. Besides, epidemic prevention and control in the real world rely on combinations of multiple policies while the existing models generally focus on developing a single policy at a time. The cooperative relationships between policies are ignored which further reduces the practicality of previous methods.

In this article, we leverage the idea of "Game and Reference" in chess players' training to tackle the above issues as epidemic policy-making and chess playing are both decision-making tasks without ground truth. It is known that the growth of a chess player requires playing games against experts to learn the way to win and referring to chess manuals to learn the playing experience in specific situations. Following this idea, we propose a brand new epidemic policy-making model named \textbf{P}olicy \textbf{C}ombination \textbf{S}ynthesis (PCS) consists of a policy generator and an adaptive multi-task learning network. In particular, the multi-task learning network is made up of an adversarial module, a contrast module, and an evaluator. The evaluator assesses the comprehensive effect of policies which includes their epidemic containment effect and economic impact. Based on the evaluator's outputs, the adversarial and contrast modules apply adaptive learning to local and global historical policies respectively. Specifically, the adversarial module plays a game between the policy generator and the discriminator. By increasing the difficulty of fooling the discriminator for the policy generator with the comprehensive effect of real policies, the policy generator is expected to learn the decision-making style of humans and avoid extreme decisions. On the other hand, to minimize the influence of subjectivity and cognitive limitations in historical policies, the contrast module forms the sample pairs of contrastive learning for the policy generator with globally retrieved real policies towards similar epidemic situations. The contrastive learning is adjusted by the comprehensive effect of real policies and the similarity between the epidemic situations the two policies oriented. In this way, the policy generator is expected to have more chance to draw on good artificial experience than referring to only local historical data. 
In addition, unlike the existing studies that only output the intensity of a single policy, the PCS model generates the intensity vector of all the given policies which fully takes the cooperative effect of policies into consideration.

Through extensive experiments on real-world data of the COVID-19 epidemic, the proposed model is proved to be more effective than the previous ones. The main contributions of this article are summarized as follows:
\begin{itemize}
    \item We propose a new model PCS for epidemic prevention and control policy-making which outputs the intensities of multiple policies at a time. To our best knowledge, this is the first study on multiple epidemic policy-making.
    \item We design an adaptive multi-task learning network based on the idea of "game and reference" to achieve efficient learning from historical policies. The learning network coordinates an adversarial module to prevent extreme decisions and a contrast module to overcome the subjectivity and recognition limitations of manual decision data.
    \item We enable adaptive adversarial learning and contrastive learning for the proposed model with the help of a multi-objective evaluator. By controlling the learning extent to the historical policies based on their comprehensive effect, the proposed model can keep drawing on useful experiences from historical policies.
\end{itemize}

\section{Related Works}

\subsection{Policy-Making based on Epidemic Prediction}
For the task of epidemic decision, one of the most important foundations is the model's capability on epidemic prediction. Based on models with high epidemic prediction performance, early epidemic policy-making studies usually consider simple policies such as assessing risk level, conducting nucleic acid testing and reassigning medical supplies et al \cite{9705122}\cite{10.1371/journal.pone.0236310}. The most significant commonality of these studies is that all the policies can be easily determined based on the prediction of future epidemics. Thus, in this stage, the focus of epidemic decision-making is actually the design of epidemic prediction model. Epidemic prediction approaches that are commonly used for epidemic policy-making can be classified into three paradigms: the epidemic model-based methods, the sequential model-based methods and the deep neural network-based ones. Models of the first paradigm are usually the variants of the classical SEIR model which infers the daily epidemic through a group of different equations \cite{ANNAS2020110072}\cite{he2020seir}. These models are good at reproducing the evolution processes of epidemics, making them very helpful for the formulation of short-term policies. The second paradigm of epidemic prediction models generally targets at epidemic trend forecasting with the help of classical sequential prediction models including MLR, LR, ARIMA et al \cite{info11090454}\cite{HERNANDEZMATAMOROS2020106610}. The estimation of epidemic trends from these models enables the policy programming in a long future period which is significant for epidemic prevention assignment of government \cite{10.1145/3485447.3512139}\cite{9613774}. The last paradigm of models applies more various data for epidemic prediction and leverages temporal neural networks to handle the non-linear or uncertain relationships between features \cite{KARA2021115153}\cite{9256562}. Though achieved initial success, early epidemic policy-making models are still of low practicality as they can only support the decision-making of limited simple policies. policies that can not be directly inferred from single epidemic data (daily new cases, accumulative cases et al) can hardly be made by these models.

\subsection{Policy-Optimizing based on Effect Evaluation}
Policy optimization through policy effect evaluation can not make new policies for human decision-makers. Instead, they provide suggestions for policy-making under certain scenarios \cite{10.1145/3394486.3412863}\cite{10.1145/3394486.3412860}. The task of policy effect evaluation is to predict the changes in future epidemics under given policies. Therefore, compared to the epidemic prediction which focuses on modeling the temporal or spatial-temporal correlation between regional epidemics, policy effect evaluation models usually consider more disturbance effect of external interventions to the epidemic trend. Due to their expertise in feature analysis, mathematical prototypes such as the Gaussian process or Markov Chain model are usually adopted to develop policy effect evaluation approaches \cite{NEURIPS2020_79a3308b}. Based on these models, epidemic policy-making becomes simple: just select the policy with the best effect evaluation. In this way, more types of policies, especially those that can not be directly mapped from epidemic prediction results, are able to be determined by the model. policies that could be determined by this paradigm of decision models include travel restrictions \cite{chinazzi2020effect}, gathering restrictions\cite{BOYER2022100620}, and vaccination \cite{10.1145/3534678.3542673}. Though becoming able to provide valuable support to epidemic policy making, the biggest problem of effect evaluation-based epidemic policy-optimizing is that the outputs of these models generally lack of foresight. These models can only tell people what needs to be noted or what might be more proper with the given epidemic situation, which can hardly replace true policy-making. Besides, suggestions provided by the existing effect evaluation-based methods failed to learn the characteristics of manual policy-making, therefore easily appear to be extreme for human decision-makers.

\section{Method}

\subsection{Problem Statement}
The task of epidemic policies-making is to determine the intensity of certain types of policies $A_{i,t}=\{A_{i,1,t+1},A_{i,2,t+1},...,A_{i,M,t+1}\}$ for region $i$ during the $(t+1)$th time slot, based on the historical epidemic data of the region $X_{i}=\{x_{i,t-l}, x_{i,t-l+1},...,x_{i,t}\}$ and the related social and economic data. Among these, epidemic data of each time slot $x_{i,r}$ is a vector including multiple epidemic-related features. Meanwhile, epidemic-related social data of a region is the intensities of its current policies $A_{i,t}$ and regional features $U_i$ (e.g., points of interest, traffic flows). The region's economic data is its Gini coefficient during each recent time slot. In convenience for expression, we use $B_{i,t}=\{X_{i,t},A_{i,t},G_{i,t},U_{i}\}$ to indicate the comprehensive situation of region $i$ which we call it the scenario of policy-making for the region in the $(t+1)$th time slot.
\subsection{Overall Framework}
\begin{figure*}[btp]
    \centering
    \includegraphics[width=0.87\textwidth]{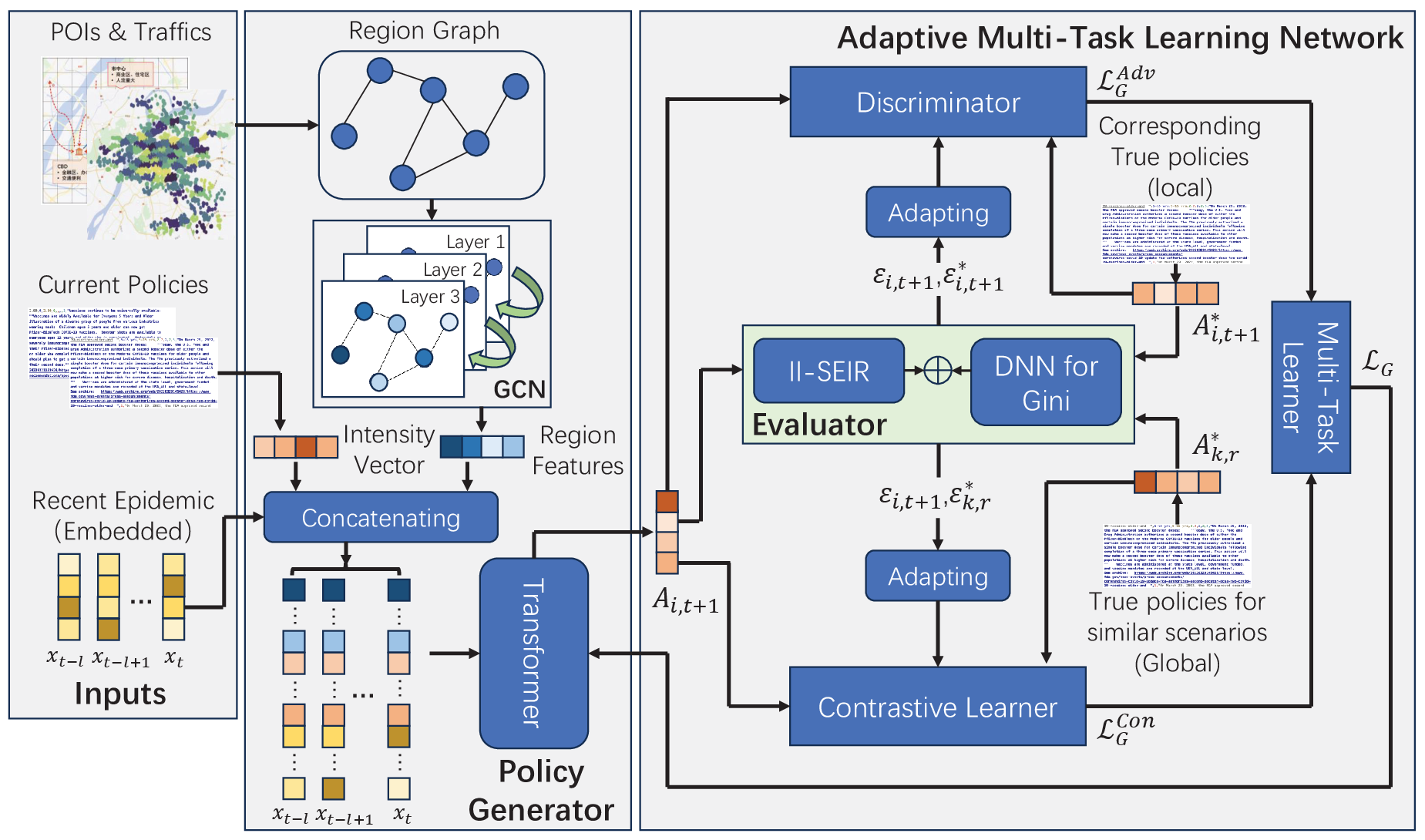}
    \caption{The framework of PCS model}
    \label{fig1}
    \vspace{-20pt}
\end{figure*}
The general framework of the proposed PCS model is displayed in Fig \ref{fig1}. As shown in the figure, the proposed model consists of a policy generator and a multi-task learning network. Given the recent epidemic data, the regional data, and the ongoing policies of a region as the input, the policy generator outputs the intensity of all types of policies in the next time slot for the region. To train the policy generator, the multi-task learning network first uses the evaluator to assess the comprehensive effectiveness of a policy and accordingly provides the policy an evaluation score. Guided by the evaluation scores, the adversarial module and the contrast module are employed to perform adaptive training on the policy generator simultaneously.
The training of the two modules is coordinated by a specially designed multi-task objective function.

\subsection{Policy Generator}
Our PCS model has no strict restrictions on the design of the policy generator. In this article, considering that the input data includes temporal data (recent epidemic of the given region and GDP records) and non-temporal data (points of interest and traffic flows), we design the policy generator as the combination of a GCN and a Transformer.

As shown in the left part of Fig \ref{fig1}, a feature graph of the points of interest $Graph_{i,t}=\{\Pi_{i},\Xi_{i,t}\}$, including their functional attributes and traffic flows between them are sent into the GCN. Among $Graph_{i,t}$, $Pi_{i}=\{\pi_{i,1}, \pi_{i,2},...,\pi_{i,n_i}\}$ indicates the attributes of all the points of interest in region $i$ and each $\pi_{i,j}$ is a functional attribute of the $j$th point of interest. Meanwhile, in the other component of $G_{i}$, $\Xi_{i,t}=\{\xi_{i,1,t},\xi_{i,2,t},...,\xi_{i,n_i,t}\}$, each $\xi_{i,j,t}$ is a vector stores the average traffic flows between $j$ and other region $i$'s points of interest in recent time.

Then we apply a three-layered GCN on $G_{i}$ to refine the regional features. The GCN applied in this article follows the standard forward transmission design which is formulated as follows:
\vspace{-7pt}
\begin{equation}
    \vspace{-7pt}
    \begin{split}
        Z_{i,t}&=softmax(\hat{C}ReLU(\hat{C}\Xi_{i,t}W^{0})W^{1}).
    \end{split}
    \label{eq1}
\end{equation}
In (\ref{eq1}), $\hat{C}$ is the sum of adjacency matrix $A$ and the identity matrix $I$. The adjacency matrix $A$ is the binary representation of $\Xi_{i,t}$, in which, $c_{i,j}=1$ for any $\xi_{i,j,t}>0$. Here, we only use three GCN layers because it has been discovered that GCN with too many layers will easily lead to an over-smooth problem \cite{NEURIPS2020_a6b964c0}.

On the other hand, the epidemic feature vector, $x_{i,r}=\{\Delta{I}_{i,r}, I_{i,r},R_{i,r}\}$, contains the number of new cases $\Delta{I}_{i,r}$, the number of accumulative cases $I_{i,r}$ and the number of recovered cases $R_{i,r}$ of region $i$ during the $r$th time slot. 
Considering the non-linear relationships between multiple features, we intend to estimate the optimal policy intensity based on the regional epidemic through a transformer \cite{NIPS2017_3f5ee243}. The input of the transformer, $X_{i,t}^{e}$, is made up by concatenating $X_{i,t}$ with $Z_{i,t}$. Then based on $X_{i,t}^{e}$, the transformer generates the intensity vector of policies for region $i$ in the $(t+1)$th time slot:
\vspace{-5pt}
\begin{equation}
    \vspace{-5pt}
    \begin{split}
        A_{i,t+1}&=Transformer(X_{i,t}^{e};\Theta). \\
    \end{split}
    \label{eq2}
\end{equation}
Where $\Theta$ is the parameter set of the transformer. Notice that (\ref{eq2}) is essentially a sequential prediction task, hence any sequential prediction method that good at processing non-linear feature relationships can replace the transformer here.

\subsection{Adaptive Multi-Task Learning Network}
The adaptive multi-task learning of the PCS model follows one principle: always let the model learn from better historical policies than it does. To do so, we design a multi-objective evaluator to assess the effectiveness of policies and thereby drive the adaptive learning of an adversarial module and a contrast module. Details of the framework are displayed in the following subsections.

\subsubsection{Multi-Objective Policy Evaluator}
The multi-objective evaluator takes both the epidemic containment effect and the economic impact into consideration in the assessment of a policy's influence. Such evaluator can be formulated as:
\vspace{-5pt}
\begin{equation}
    \vspace{-5pt}
    \begin{split}
        \varepsilon_{i,t+1}&=\Omega(\Delta{I}_{i,t+1},G_{i,t+1};\Psi) \\
        G_{i,t+1}&=\Gamma(I_{i,t+1},E_{i,t+1},R_{i,t+1};\varphi). \\
    \end{split}
    \label{eq3}
    \vspace{-10pt}
\end{equation}
In (\ref{eq3}), $\Delta{I}_{i,t+1}$ denotes the estimated new case reduction caused by the policies, which is predicted by our specially designed II-SEIR module (section \ref{sec3-1}). $\Omega(\cdot)$ and $\Gamma(\cdot;\varphi)$ are both deep neural networks used to compute the policy score $\varepsilon_{i,t+1}$. $G_{i,t}$ indicates the Gini coefficient which is a widely-used economic index. 

Notice that in theory, (\ref{eq3}) can largely prevent the model from extreme decisions because of its multi-objective scoring. But since there exists huge difference between the value ranges of $\Delta{I}_{i,t+1}$ and $G_{i,t+1}$ ($\Delta{I}_{i,t+1}\in[0,\infty]$ and $G_{i,t+1}\in[0,1]$), it is hard to balance the weight of the two objectives in model training, making (\ref{eq3}) fail to have the expected constraint power on model. Thus, it is important to let the model learn more about the characteristics and the experience of manual decision-making, so as to develop effective and reasonable policies. Such issue prompts us to design the adversarial module and the contrast module in section \ref{sec3-2} and \ref{sec3-3} respectively.

\begin{figure*}[btp]
    \vspace{-10pt}
    \centering
    \subfloat[SEIR model]{
    \includegraphics[width=2.35in]{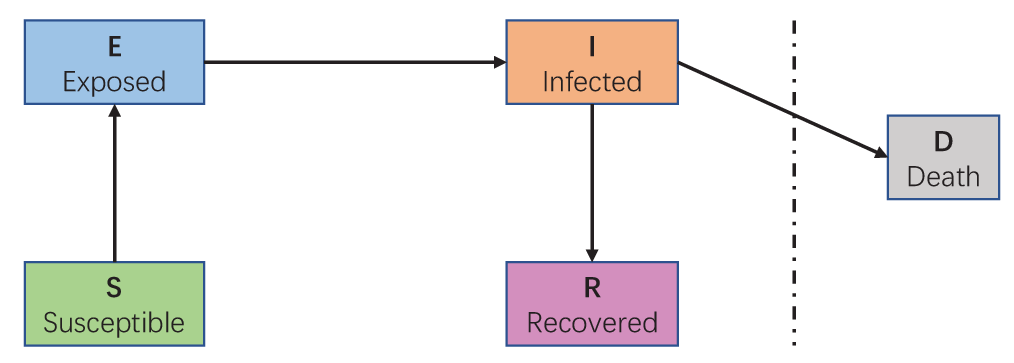}
    }
    \subfloat[II-SEIR model]{
    \includegraphics[width=2.35in]{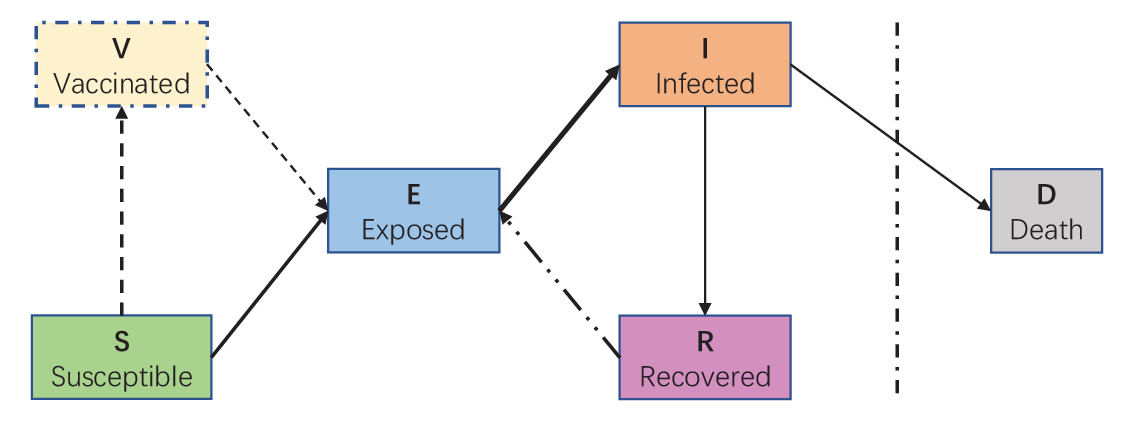}
    }
    \caption{Comparison of SEIR and the proposed II-SEIR}
    \label{fig2}
    \vspace{-20pt}
\end{figure*}

\subsubsection{The II-SEIR epidemic prediction module}\label{sec3-1}
The \textbf{I}mmuity \textbf{I}ntervened \textbf{SEIR} (II-SEIR) module is the core of multi-objective policy evaluator in (\ref{eq3}), it provides the prediction of future new cases for both the policy score $\varepsilon_{i,t+1}$ and the GDP estimation $G_{i,t+1}$. This model is derived from the most widely adopted epidemic model SEIR which tackles two issues ignored by SEIR and its existing variants: the impact of policies and the intervention of immunity.

In SEIR model, people in both \textbf{S}usceptible (S) and \textbf{R}ecovery (R) states transfer to \textbf{E}xposed (E) group with a fixed probability based on the epidemic or the strain. This assumption only applies to the early stage of the epidemic when there is no vaccination or antibody against new strains. With the evolution of the epidemic, antibodies from vaccines and recovered people will contain the transformation. On the other hand, policies especially the NPIs have a significant containing effect on the spread of the virus. Thus, by taking the above two factors into consideration, we have the II-SEIR module shown in Fig \ref{fig3}.

From Fig \ref{fig3}, it can be found that besides the four states in SEIR, the II-SEIR module contains an additional state \textbf{V}accinated (V) to indicate the people who have been vaccinated. Accordingly, extra state transfer paths are constructed between state V, S, R and E. For current epidemic strain $q$, the state transfer rules of the II-SEIR module can be formulated as follows:
\vspace{-5pt}
\begin{equation}
    \vspace{-5pt}
    \begin{split}
        &\Delta{S}_{i,t}=-\psi_{i,t}I_{i,t}S_{i,t}[(1-p_{i,t}^{v})+p_{i,t}^{v}\sum_{q}{\phi_{i,t}^{q}(1-\delta_{q})}] \\
        &\Delta{E_{i,t}}=\psi_{i,t}I_{i,t}S_{i,t}+R_{i,t}\sum_{s}{\rho_{i,t}^{s}(1-\delta_{sq})}-\beta{E_{i,t}} \\
        &\Delta{I_{i,t}}=\beta{E_{i,t}}-\mu{I_{i,t}} \\
        &\Delta{R}=\mu{I_{i,t}} \\
        &\Delta{V}=S_{i,t}p_{i,t}^{v}. \\
    \end{split}
    \label{eq4}
\end{equation}
In (\ref{eq4}), $p_{i,t}^{v}$ is the proportion of the exposed people who have been vaccinated during the $t$th time slot in region $i$, $\delta_{q}$ and $\delta_{sq}$ are the immune coefficient of the vaccine and the strain $s$'s antibody to strain $q$. $\psi_{i,t}$ denotes the probability that an exposed person being infected during the $t$th time slot. Unlike the existing epidemic models set $\psi_{i,t}$ as a constant, we connect $\psi_{i,t}$ with the intensities of policies to bring in the policies' impact:
\vspace{-5pt}
\begin{equation}
    \vspace{-5pt}
    \begin{split}
        \psi_{i,t}&=\psi_{i,t-1}+F(\Delta{A}_{i,t,t-1};\theta). \\
    \end{split}
    \label{eq5}
\end{equation}
Where $F(\cdot;\theta)$ is a deep neural network with parameter set $\theta$, $\Delta{A}_{i,t,t-1}$ denotes the variation of policies' intensities from the $(t-1)$th time slot to the $t$th time slot. The reason why we use $\Delta{A}_{i,t,t-1}$ instead of $A_{i,t}$ in (\ref{eq5}) is that the II-SEIR inherits the difference modeling framework of SEIR, in which, $A_{i,t}$ is only applied in the computation of the first time slot.

\subsection{Adversarial Module}\label{sec3-2}
The adversarial module enables the proposed model to make more human-liked policies through adversarial learning. Existing adversarial learning methods usually force the model output unconditionally close to the real data, so as to deceive the discriminator. Such a paradigm of adversarial learning is effective when there exists reliable ground truth. Unfortunately, due to subjectivity and cognitive limitations, historical policies are not always good enough to be treated as the ground truth. Thus, the adversarial module in our model is equipped with an adaptive control mechanism to adjust the model's learning according to the quality of real data. Based on the evaluation score $\varepsilon_{i,t+1}^{*}$ of the corresponding real policy, we have the objective functions of the discriminator and the generator in our adversarial module as follows:
\vspace{-5pt}
\begin{equation}
    \vspace{-5pt}
    \begin{split}
        \mathop{\arg\max}&\log{D(A_{i,t+1}^{*})}+\log{[1-\sigma(\varepsilon_{i,t+1},\varepsilon_{i,t+1}^{*})D(A_{i,t+1})]} \\
        \mathop{\arg\min}&\log{[1-\sigma(\varepsilon_{i,t+1},\varepsilon_{i,t+1}^{*})D(A_{i,t+1})]}. \\
    \end{split}
    \label{eq6}
\end{equation}
In (\ref{eq6}), $\sigma(\cdot)$ is an activation function used to generate the control parameter based on the comprehensive scores of each pair of policies. The adaptive rule in (\ref{eq6}) is: when the score of real policy is lower than that of the model output, $\sigma(\varepsilon_{i,t+1},\varepsilon_{i,t+1}^{*})$ increases to help the generator to deceive the discriminator. Relatively, when the real policy has a higher score, we make it more easy for the discriminator to distinguish the output policy from the real one. Such a rule prevents the model from learning low-quality real data and encourages it to learn more from data with higher quality. Based on (\ref{eq6}), the loss function for the policy generator is:
\vspace{-5pt}
\begin{equation}
    \vspace{-5pt}
    \begin{split}
        \mathcal{L}_{G}^{Adv}&=N^{-1}\sum_{i=1}^{N}{\log{D(A_{i,t+1}^{*})}-\log{[1-\sigma(\varepsilon_{i,t+1},\varepsilon_{i,t+1}^{*})D(A_{i,t+1})]}}. \\
    \end{split}
    \label{eq7}
\end{equation}
Since there is little for the generator to learn from real data when the gap between $\varepsilon_{i,t+1}$ and $\varepsilon_{i,t+1}^{*}$ is too small or too large, we'd like to let the change of learning restriction be more significant during the rest section of $\varepsilon_{i,t+1}-\varepsilon_{i,t+1}^{*}$. Hence, we choose the hyperbolic tangent function $tanh(\cdot)$ as the $\sigma(\cdot)$ in (\ref{eq6}) and (\ref{eq7}).

\subsection{Contrast Module}\label{sec3-3}
As mentioned before, the subjectivity and cognitive limitations make local historical policies not always optimal against the corresponding scenarios. To overcome this issue, we incorporate contrastive learning to let the model draw on experience from high-scored historical policies of all the regions. Similar to the adversarial module in \ref{sec3-2}, the contrast module is also adaptive. However, since the historical policies in sample pairs may be from different regions at different times, the contrastive learning of the proposed model has an additional controlling factor besides the policies' comprehensive scores: the similarity between the scenarios of the two policies. Based on the classic inforNCE loss of contrastive learning, the adaptive infoNCE loss of our contrast module is:
\begin{equation}
    \begin{split}
        l_{i,k}^{C}&= \frac{H(\varepsilon_{i,t+1},\varepsilon_{k,r+1}^{*},B_{i,t},B_{k,r})}{\sum_{j|j\neq{k}}{exp[sim(B_{i,t},B_{j,r})/\tau]}+H(B_{i,t},B_{k,r})} \\
        H(&\varepsilon_{i,t+1},\varepsilon_{k,r+1}^{*},B_{i,t},B_{k,r})= [1+\sigma(\varepsilon_{i,t+1},\varepsilon_{k,r+1}^{*})]e^{sim(B_{i,t},B_{k,r})/\tau}. \\
    \end{split}
    \label{eq8}
\end{equation}
Where $\sigma(\cdot)$ is an activation function to generate policy score-based adaptive coefficient, $sim(B_{i,t},B_{k,r})$ denotes the cosine similarity between the scenarios $B_{i,t}$ and $B_{k,r}$ of region $i$ and region $k$ respectively. Then the contrastive loss for the generator will be:
\vspace{-5pt}
\begin{equation}
    \vspace{-5pt}
    \begin{split}
        \mathcal{L}_{G}^{Con}&=N^{-1}K^{-1}\sum_{i=1}^{N}{\sum_{k=1|k\neq{i}}^{K}{l_{i,k}^{C}}}. \\
    \end{split}
    \label{eq9}
\end{equation}
In (\ref{eq9}), $K$ denotes the number of sample pairs in each round of contrastive learning. To ensure that the model always learns from high-quality policies towards similar scenarios, we set a filtering threshold:
\vspace{-5pt}
\begin{equation}
    \vspace{-5pt}
    \begin{split}
        z_{i,k}&=\mathop{max}\{0,\mathop{sgn}(sim(B_{i,t},B_{k,r})-Z_{th})\}. \\
    \end{split}
    \label{eq10}
\end{equation}
In (\ref{eq10}), $z_{i,k}$ is a binary indicator whose value will be $1$ if the scenario $B_{k,r}$ is close enough to $B_{i,t}$ to form the sample pairs with model outputs. Otherwise, the value of $z_{i,k}$ will be $0$. $Z_{th}$ is the scenario similarity threshold for real policy selection. Only historical polices that make $z_{i,k}=1$ will be selected for the contrastive learning.

\subsection{Multi-Task Training}
To coordinate the two heterogeneous modules in \ref{sec3-2} and \ref{sec3-3} during the training of the proposed model, one of the most widely adopted and mature way is multi-task learning. Typical multi-task learning constructs a linear combination of multiple loss functions to train the model under different sub-tasks simultaneously \cite{zhang2018overview}. Following this paradigm, the overall loss function for the training of the policy generator is:
\vspace{-5pt}
\begin{equation}
    \vspace{-5pt}
    \begin{split}
        \mathcal{L}_{G}&=\lambda\cdot{\mathcal{L}_{G}^{Con}}+(1-\lambda)\cdot{\mathcal{L}_{G}^{Adv}}. \\
    \end{split}
    \label{eq11}
\end{equation}
Where $\lambda\in[0,1]$ in (\ref{eq11}) is a control parameter to balance the weight of the two sub-tasks during the model training. As a start of our epidemic policy-making study, we set $\lambda$ in (\ref{eq11}) as a manual turned constant. But it is feasible to upgrade it to a learnable parameter in the future.

\section{Experiments}
The experiments reported in this article is expected to answer the following \textbf{R}esearch \textbf{Q}uestions (RQs):
\begin{itemize}
    \item RQ1: How does the PCS model perform compared to the previous studies?
    \item RQ2: Is the multi-objective based evaluator precise enough to support efficient policy-making?
    \item RQ3: Can the adversarial module prevent the proposed model from making extreme decisions?
    \item RQ4: How much can the contrast module help the proposed model in its performance?
\end{itemize}

\subsection{Data Preparation}
In this article, we adopt the real-world COVID-19 epidemic data of the US from 2020 to 2022. The immunity data of vaccines and antibodies during the corresponding period are from the official statistical data of the CDC. Besides, since the PCS model and some baselines require multi-domain information, we also collect the data of policies and GDP records corresponding to the above epidemic data. To reduce the negative influence of low-quality samples and noise in the original data set, we filter out the policies without the implementation time and the policies with too short effective time. Meanwhile, we also remove the regions whose epidemic is too mild as these regions usually have few policies implied. The final data applied to the experiments includes the epidemic, policies, and GDP data of all the $50$ states, last from $2020/1/1$ to $2022/10/31$. Some details of the experimental data are shown in Table \ref{table 2}.
\begin{table}[]
    \vspace{-25pt}
    \centering
    \caption{Statistic details of experimental data}
    \begin{tabular}{|c|c|}\hline
         Number of states & 50 \\ 
         Number of policies & 2423 \\
         Time frame & 2020/1/1 - 2021/10/31\\
         Types of policies & 4 \\
         policy intensity levels & 4 \\
         policies per state & 49 \\
         \hline
    \end{tabular}
    \label{table 2}
    \vspace{-30pt}
\end{table}

\begin{figure*}[btp]
    \vspace{-5pt}
    \centering
    \subfloat[Indiana]{
    \includegraphics[width=1.1in]{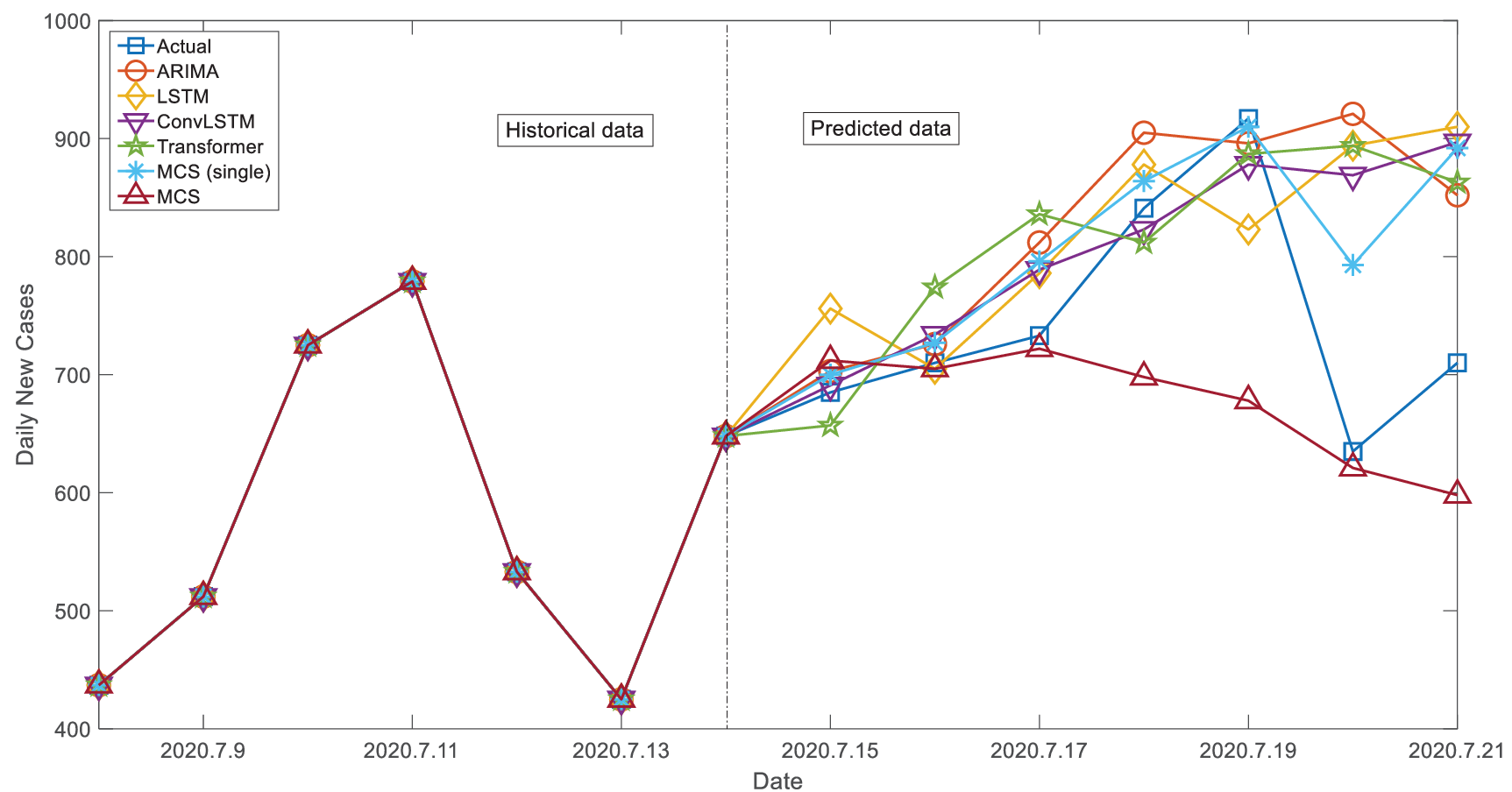}
    }
    \subfloat[Minnesota]{
    \includegraphics[width=1.1in]{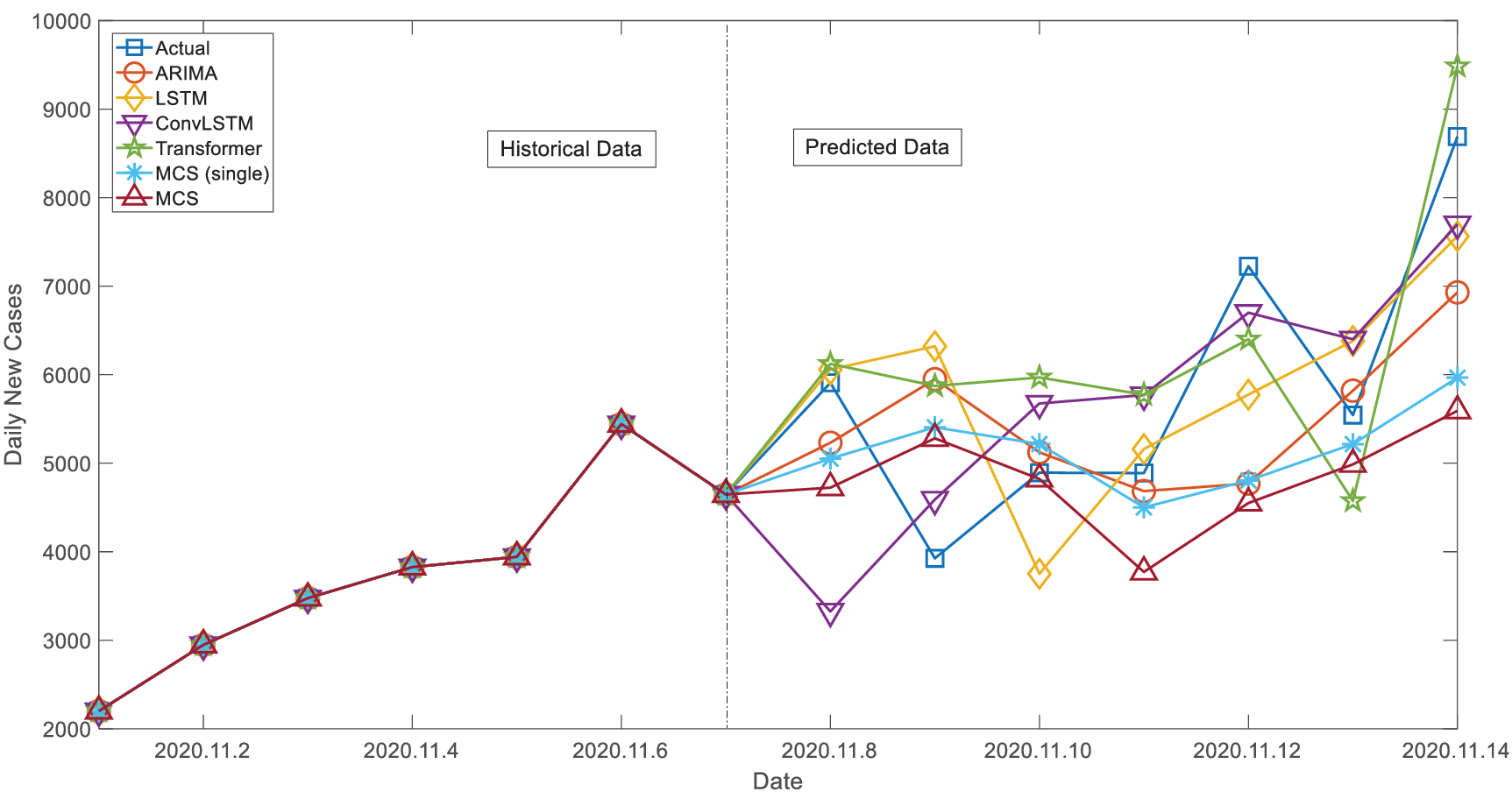}
    }
    \subfloat[Maine]{
    \includegraphics[width=1.1in]{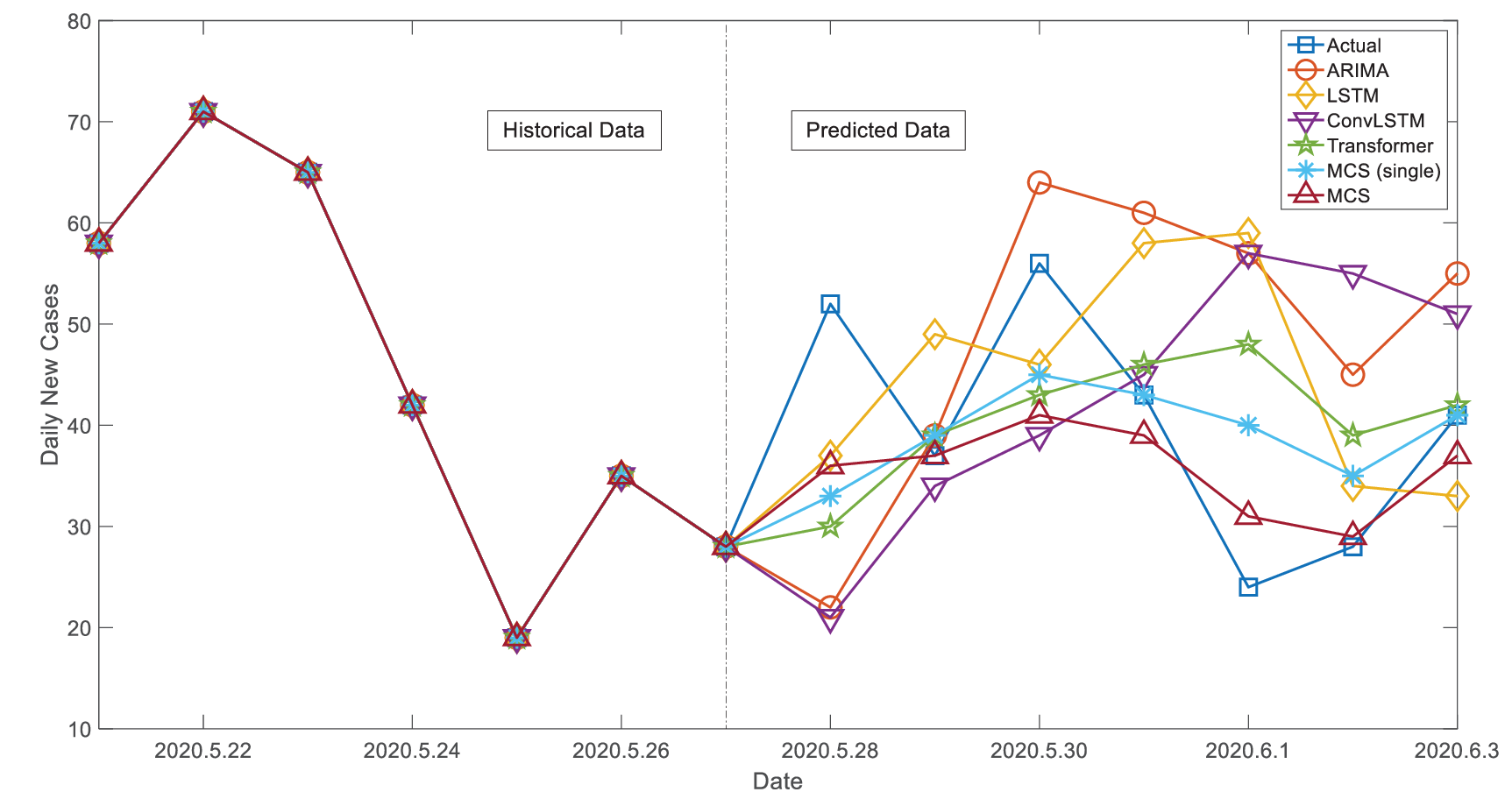}
    }
    \subfloat[Oklahoma]{
    \includegraphics[width=1.1in]{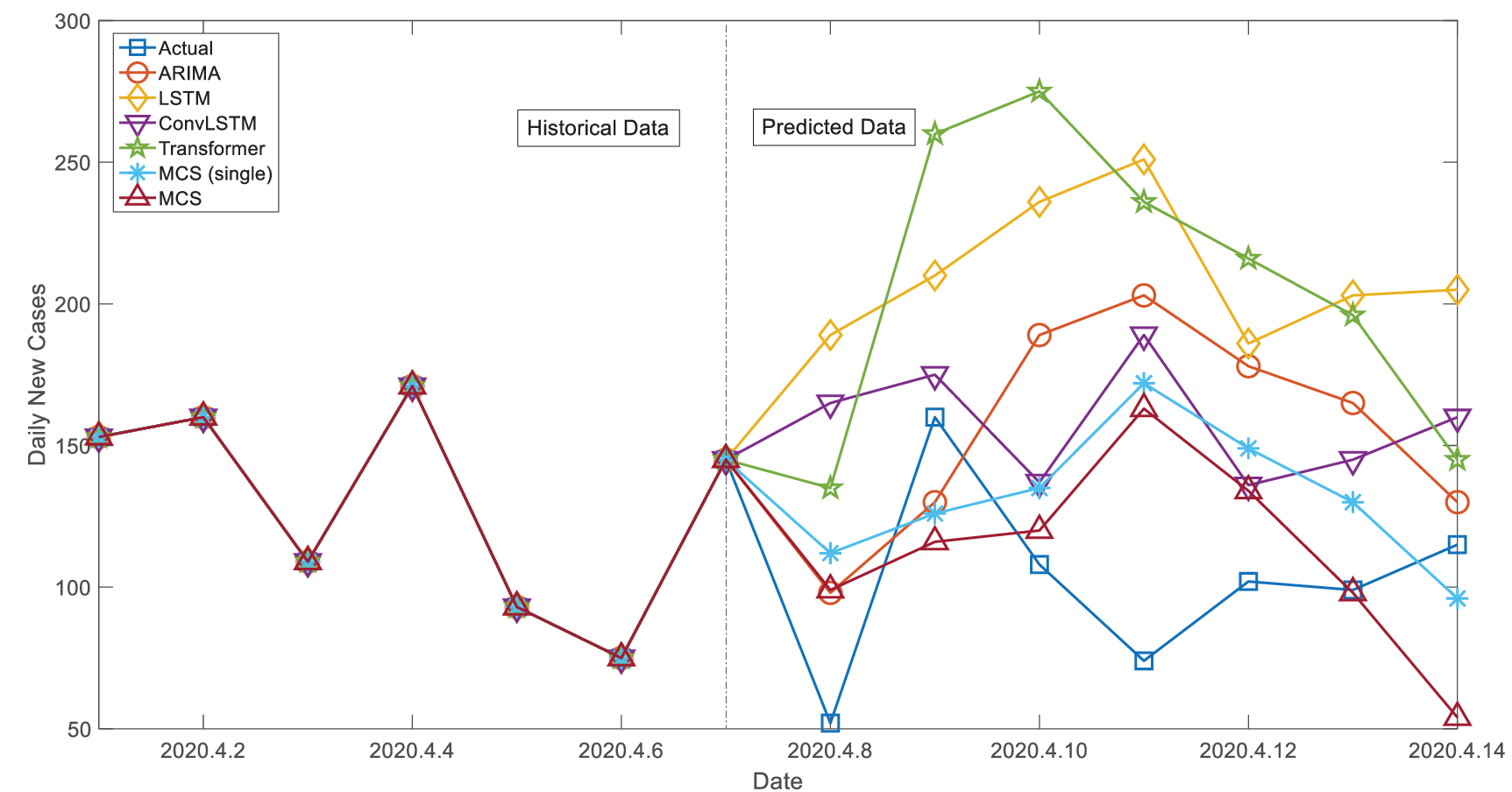}
    }
    \caption{Epidemic containment effect comparison of single policy}
    \label{fig3}
    \qquad

    \vspace{-5pt}
    \subfloat[Indiana]{
    \includegraphics[width=1.1in]{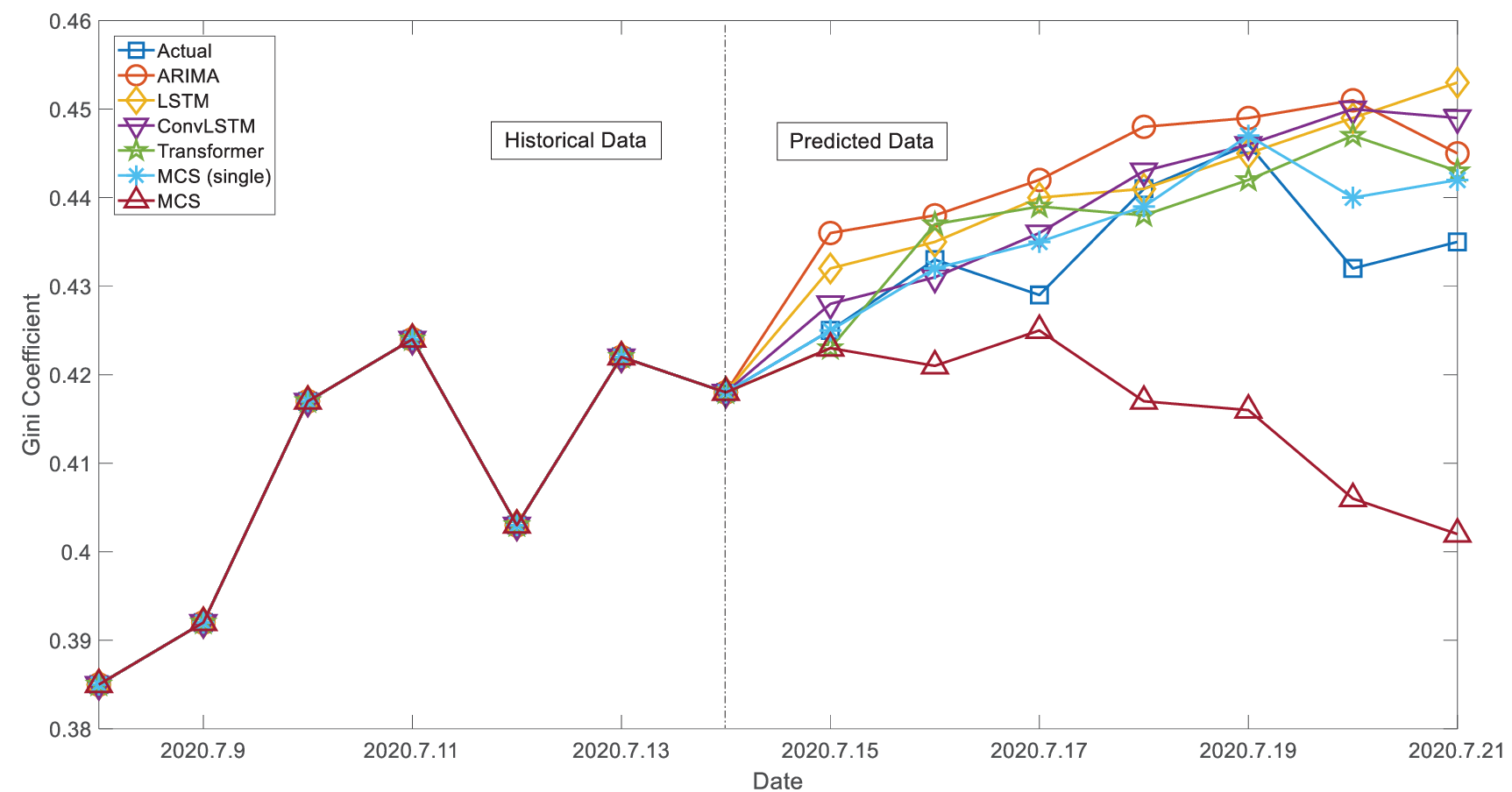}
    }
    \subfloat[Minnesota]{
    \includegraphics[width=1.1in]{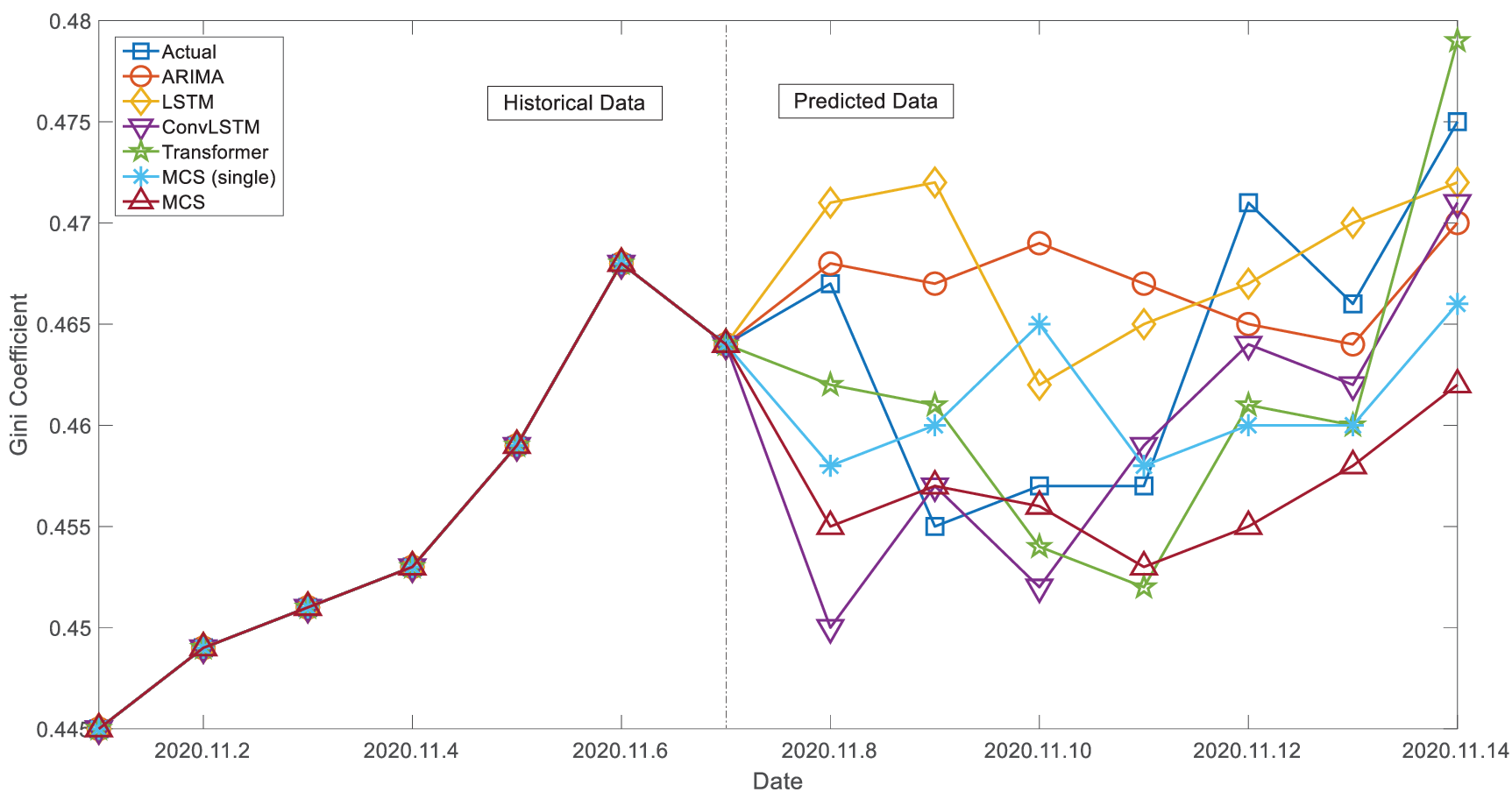}
    }
    \subfloat[Maine]{
    \includegraphics[width=1.1in]{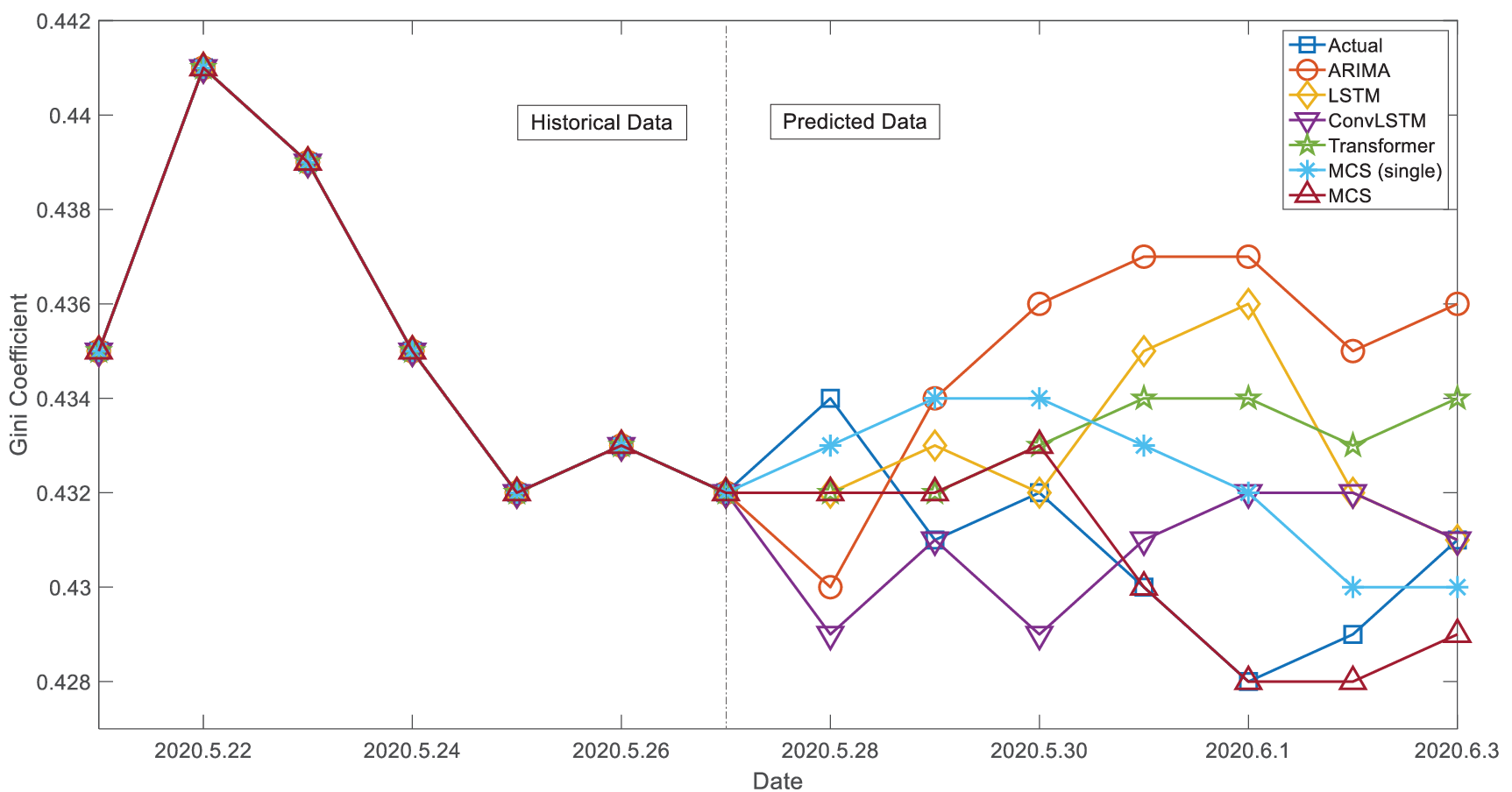}
    }
    \subfloat[Oklahoma]{
    \includegraphics[width=1.1in]{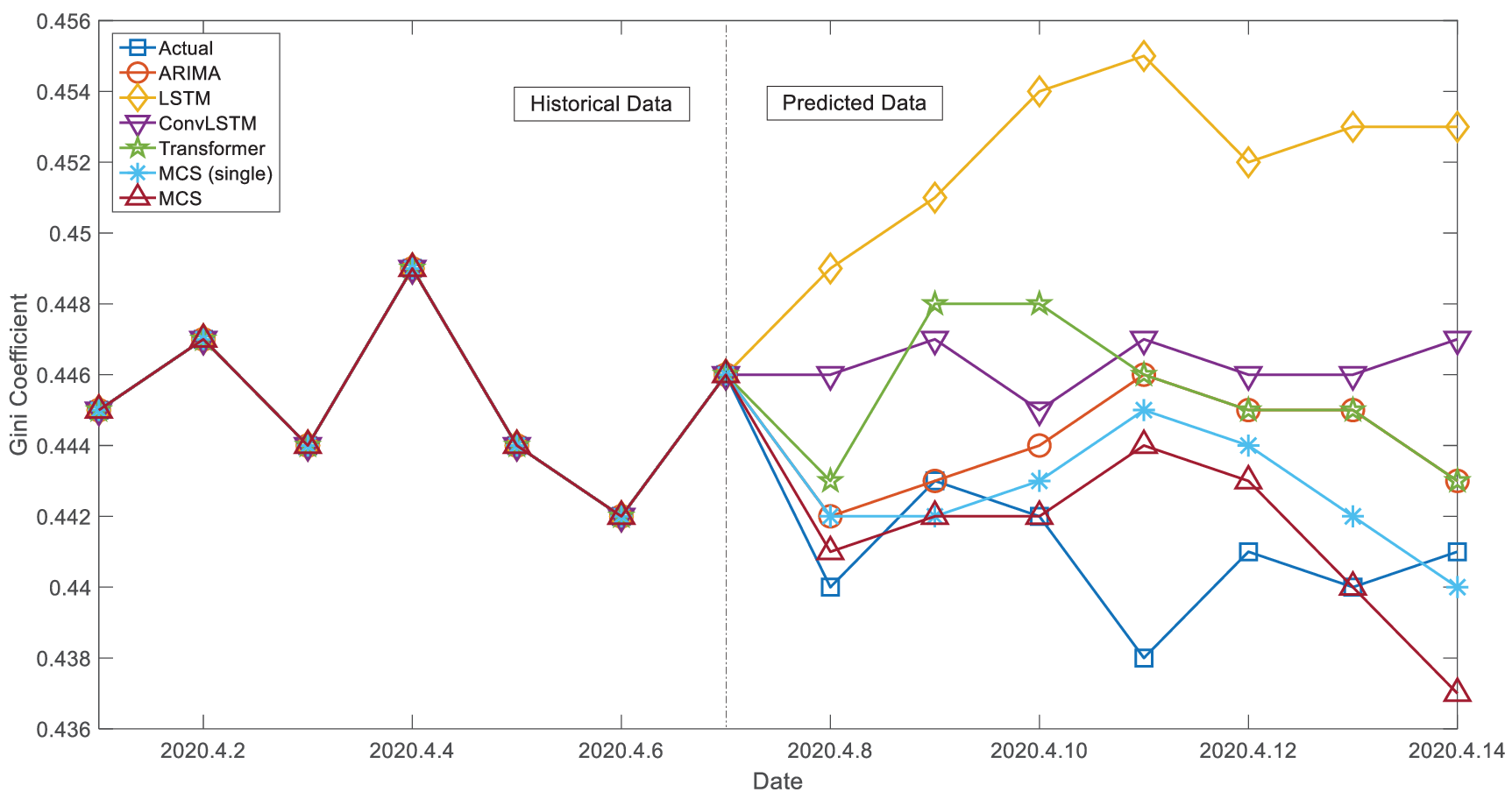}
    }
    \caption{Economic impact comparison of single policy}
    \label{fig4}
    \vspace{-20pt}
\end{figure*}

\subsection{Experiments Setting}
Since policies take time to be effective, for each given region, we set the target of all the models as the policies that going to be implied in a future period. Based on this setting, policies of a region implied in each fixed length period are treated as a sample for model training. 
Further, considering that governments of a region barely release multiple similar policies in a short period, for each policy type in a sample, we only retain the policy with the highest intensity so as to form an intensity vector for the training of the PCS model. Following the previous studies \cite{HERNANDEZMATAMOROS2020106610}\cite{info11090454}, the time frame of model input and the length of the target period is $7$ days.

There are two types of experimental results displayed in this article: the policy effect visualization and the epidemic prediction performance. For the visualization of the policy effect, we will compare the diagrams of predicted new cases and Gini coefficients with the implementation of policies from different models respectively. The epidemic prediction performance is evaluated by errors between model-estimated new cases and the real values. For this part of the evaluation, we leverage two widely adopted metrics. The first metric is the \textbf{M}ean \textbf{A}verage \textbf{E}rror (MAE) which is computed as:
\vspace{-10pt}
\begin{equation}
    \vspace{-5pt}
    \begin{split}
        MAP&=N^{-1}\cdot{\sum_{i=1}^{N}{|\hat{y}_{i}-y_{i}^{*}|}} \\
    \end{split}
\end{equation}
The MAE reflects the absolute bias of prediction results to real data. As a supplement, the second metric \textbf{R}elative \textbf{M}ean \textbf{S}quare \textbf{E}rror (RMSE) is a relative error measurement:
\vspace{-10pt}
\begin{equation}
    \vspace{-5pt}
    \begin{split}
        RMSE&=N^{-1}\cdot{\sum_{i=1}^{N}{(\frac{\hat{y}_{i}-y_{i}^{*}}{y_{i}^{*}})^{2}}} \\
    \end{split}
\end{equation}
The RMSE reflects the deviation degree of model output to the ground truth which is more meaningful than the MAE in assessing the performance of models.

For the fairness of comparison in policy effect evaluation, diagrams of the same category but different models are generated by the same model. In particular, the new case diagrams are generated by II-SEIR, and the Gini coefficient diagrams are obtained through the deep neural network $\Gamma(\cdot;\varphi)$ in (\ref{eq3})

\subsection{Baselines}
To evaluate the effectiveness of the proposed model, we choose four representative models in epidemic policy-making, including the most commonly used \textbf{LSTM} \cite{KARA2021115153} and \textbf{ARIMA} \cite{HERNANDEZMATAMOROS2020106610}, LSTM's variant \textbf{ConvLSTM} \cite{Lin_Li_Zheng_Cheng_Yuan_2020}, and \textbf{Transformer} which is the core of the proposed model's policy generator \cite{NIPS2017_3f5ee243}. Specifically, to make the comparison between the proposed model and the Transformer an extra ablation study of the whole adaptive multi-task learning network, we provide the transformer the same input as the proposed model.

\begin{figure*}[btp]
    \vspace{-5pt}
    \centering
    \subfloat[Indiana]{
    \includegraphics[width=1.1in]{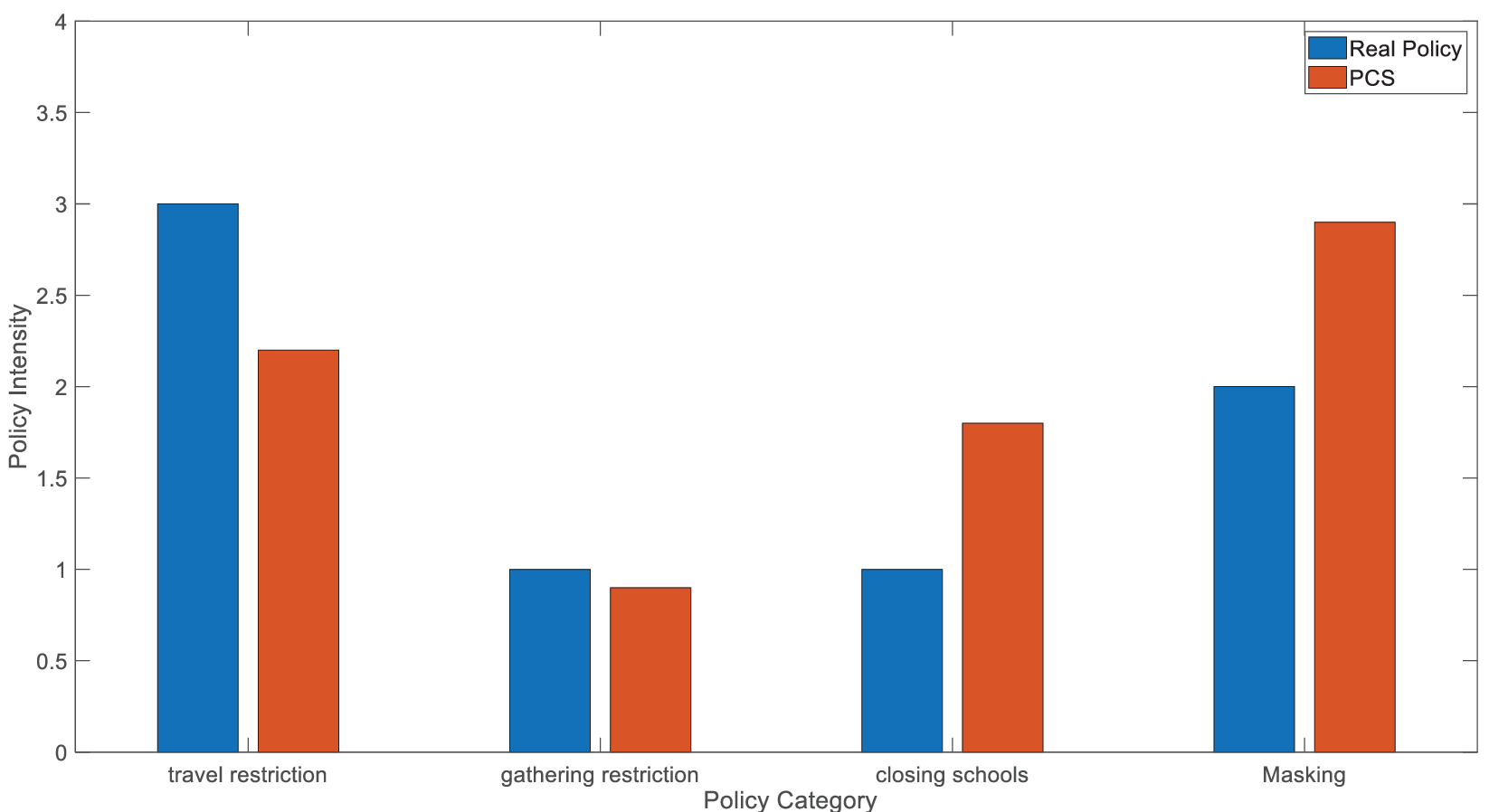}
    }
    \subfloat[Minnesota]{
    \includegraphics[width=1.1in]{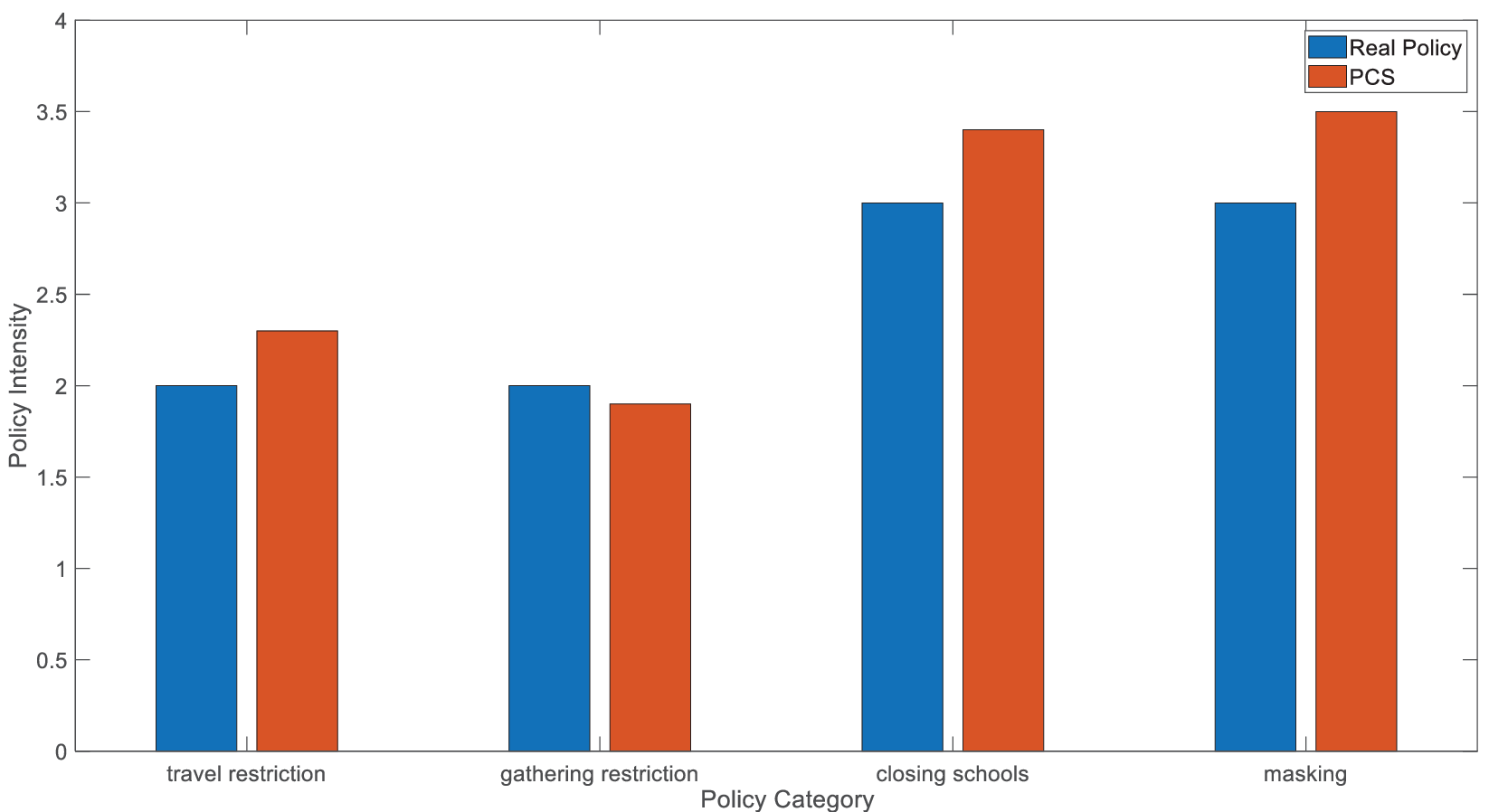}
    }
    \subfloat[Maine]{
    \includegraphics[width=1.1in]{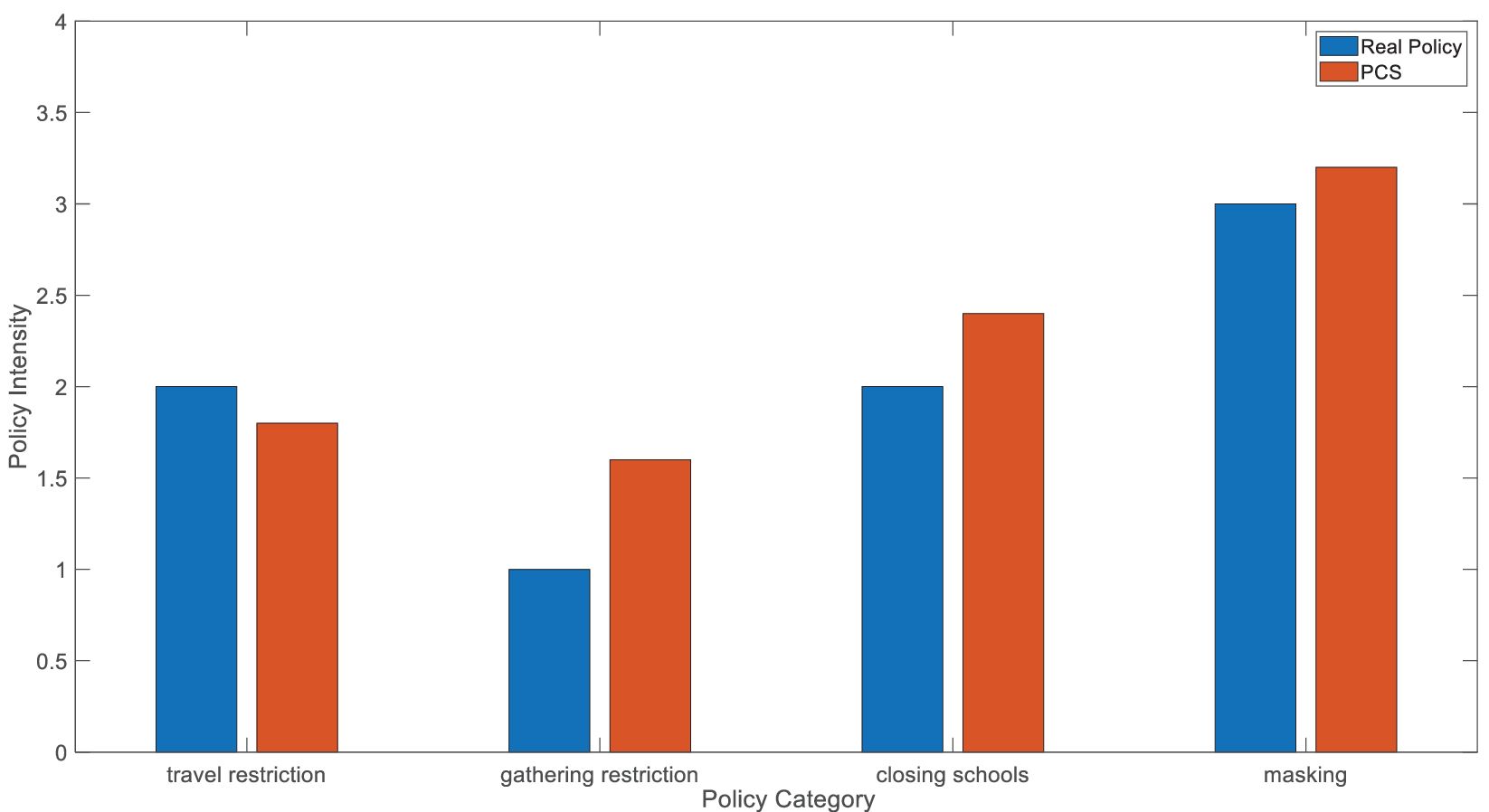}
    }
    \subfloat[Oklahoma]{
    \includegraphics[width=1.1in]{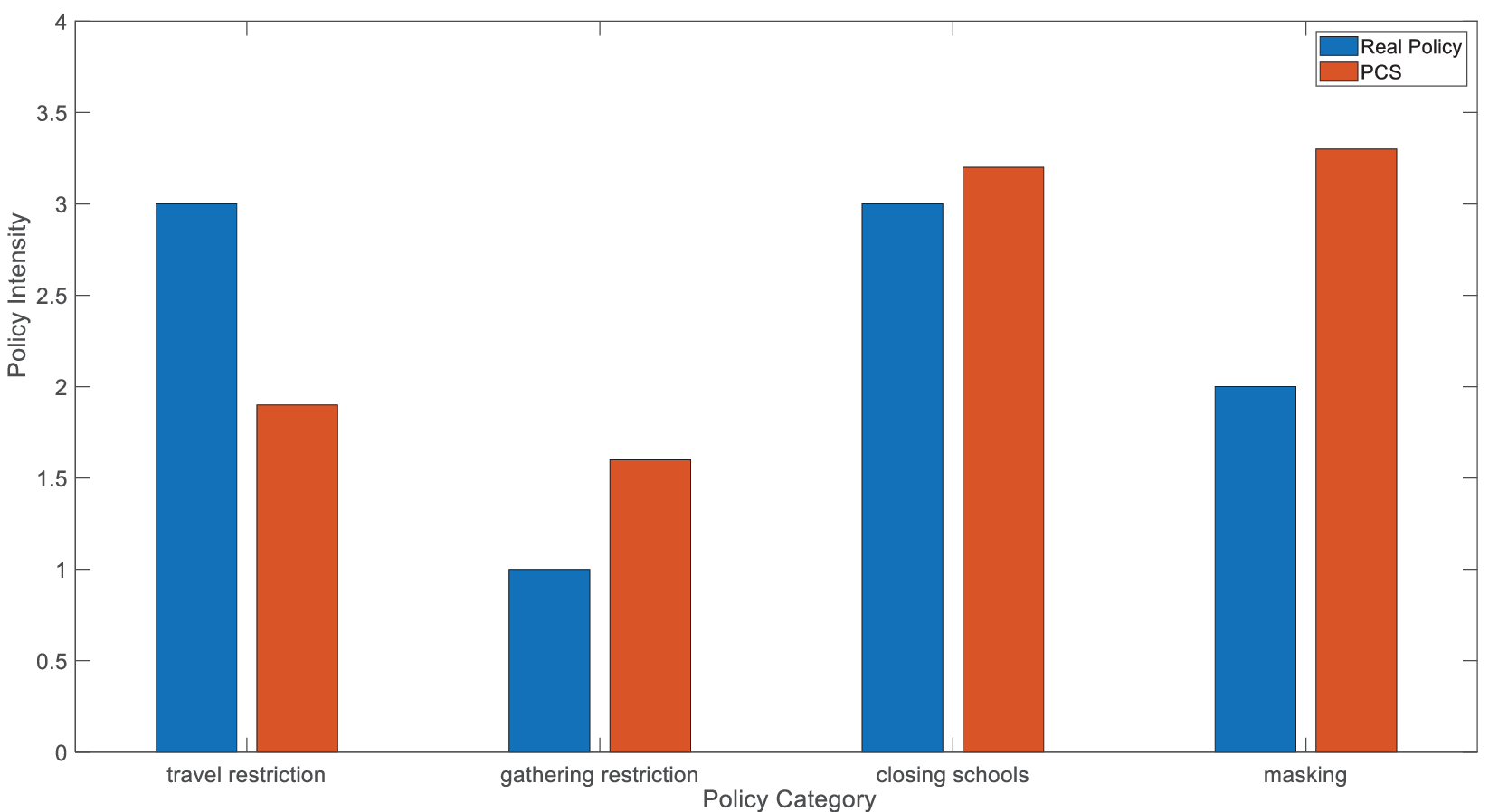}
    }
    \caption{Comparison of PCS-made policies and real policies}
    \label{fig5}
    \vspace{-20pt}
\end{figure*}

\subsection{Performance Evaluation (RQ1)}
In this section, for given policies, we compare the epidemic containment effect in terms of predicted new cases and the economic impact in terms of Gini coefficient. Considering that it is not fair to compare the effect of policy combination with that of a single policy, we add both the single policy and the policy combination made by the proposed model into the comparison. For the effect comparison of the single policy, we chose the most commonly applied "travel restriction". The results are shown in Fig \ref{fig3} and Fig \ref{fig4}.

From the \ref{fig3}, it can be found that in the comparison of a single policy's effect, the proposed model's performance is generally close to the baseline models (indicated by legend "MCS (single)"). Meanwhile, the containment effects of single policies from all the models are worse than the real data. This is because the real data is evolved under a combination of policies which brings more comprehensive epidemic containment capability. On the other hand, the epidemic containment effect of the policy combination made by the proposed model is very close to the real data and is even better in some regions (see "MCS" in Indiana and Minnesota). Meanwhile, in Fig \ref{fig4}, the Gini coefficient under the proposed model's output policy is better than those under baseline models' outputs (the optimal range for US is $0.38$ to $0.44$). Such results demonstrate that the proposed model can effectively balance the epidemic containment effect and the economic impact of policies during the decision-making process. Besides, the results in the above two figures also prove the necessity of making policy combinations rather than single policies for epidemic prevention and control.

Facing the performance of the proposed model, we are curious about why the policy combination can achieve a better epidemic containment effect and economic index simultaneously. Thus we further display the corresponding policy combinations in the four regions in Fig \ref{fig5}. From the figure, it can be found that compared to the real policies, the proposed model tends to assign higher intensity to policies of "closing schools" and "masking" and the intensities of all four types of policies are more balanced in general. Since "masking" provides the most direct protection against the virus and people in Western countries mostly hate the travel and gathering restrictions, the comprehensive effect of the proposed model's policy combination overtakes that of the real policies.

\subsection{Assessment of the Evaluator (RQ2)}
Since there is no previous study on epidemic policies-making models correlations between epidemic features and economic index like the Gini coefficient, we only compare the prediction performance of the II-SEIR module in the evaluator with baseline models in this section. Similar to most previous studies, the goal of epidemic prediction is the number of new cases during the target period. The results are shown in Table \ref{table3}.

\begin{table*}[]
    \vspace{-25pt}
    \centering
    \renewcommand\arraystretch{1.35}
    \caption{Epidemic Prediction Performance of II-SEIR}
    \begin{tabular}{cc|ccccc|c}\hline
        \multicolumn{2}{c|}{Models} & Indiana & Minnesota & Maine & Oklahoma & New York & Average \\\hline
        \multirow{2}{*}{LSTM} & MAE & 381.12 & 3040.47 & 16.58 & 38.18 & 736.59 & 2112.48 \\
         & RMSE & 0.2612 & 0.2685 & 0.1706 & 0.1417 & 0.2834 & 0.2459 \\\cline{1-8}
        \multirow{2}{*}{ARIMA} & MAE & 380.24 & 3048.95 & 16.57 & 49.80 & 736.56 & 2193.71 \\
         & RMSE & 0.2609 & 0.2702 & 0.1701 & 0.2411 & 0.2830 & 0.2653 \\\cline{1-8}
        \multirow{2}{*}{ConvLSTM} & MAE & 346.26 & 2706.16 & 15.75 & 34.26 & 669.63 & 1848.63 \\
         & RMSE & 0.2147 & 0.2127 & 0.1540 & 0.1141 & 0.2339 & 0.1884 \\\cline{1-8}
        \multirow{2}{*}{Transformer} & MAE & 347.43 & 2843.27 & 17.57 & 36.91 & 640.21 & 1867.67 \\
         & RMSE & 0.2157 & 0.2348 & 0.1916 & 0.1324 & 0.2138 & 0.1923 \\\cline{1-8}
         \multirow{2}{*}{PCS (II-SEIR)} & MAE & \textbf{260.48} & \textbf{2514.34} & \textbf{12.44} & \textbf{25.72} & \textbf{446.08} & \textbf{1300.92} \\
         & RMSE & \textbf{0.1215} & \textbf{0.1348} & \textbf{0.0961} & \textbf{0.0643} & \textbf{0.1038} & \textbf{0.0933} \\\hline
    \end{tabular}
    \label{table3}
    \vspace{-15pt}
\end{table*}

It can be seen from Table \ref{table3} that the proposed model outperforms all the baseline models with a quite obvious advantage. In particular, the MAE of the proposed model's prediction is about $30\%$ lower than the SOTA method (ConvLSTM) while the gap under RMSE is over $50\%$. In some regions, the gaps are even larger. This is largely because the proposed II-SEIR considers the immunity between the antibodies of different strains and merges the impact of policies into the epidemic modeling. Such results demonstrate that the proposed model has sufficient epidemic prediction capability to support its epidemic decision-making.

\subsection{Ablation Study on Adversarial Module (RQ3)}\label{sec4-6}
\begin{figure*}[btp]
    \vspace{0pt}
    \centering
    \subfloat[Indiana]{
    \includegraphics[width=1.1in]{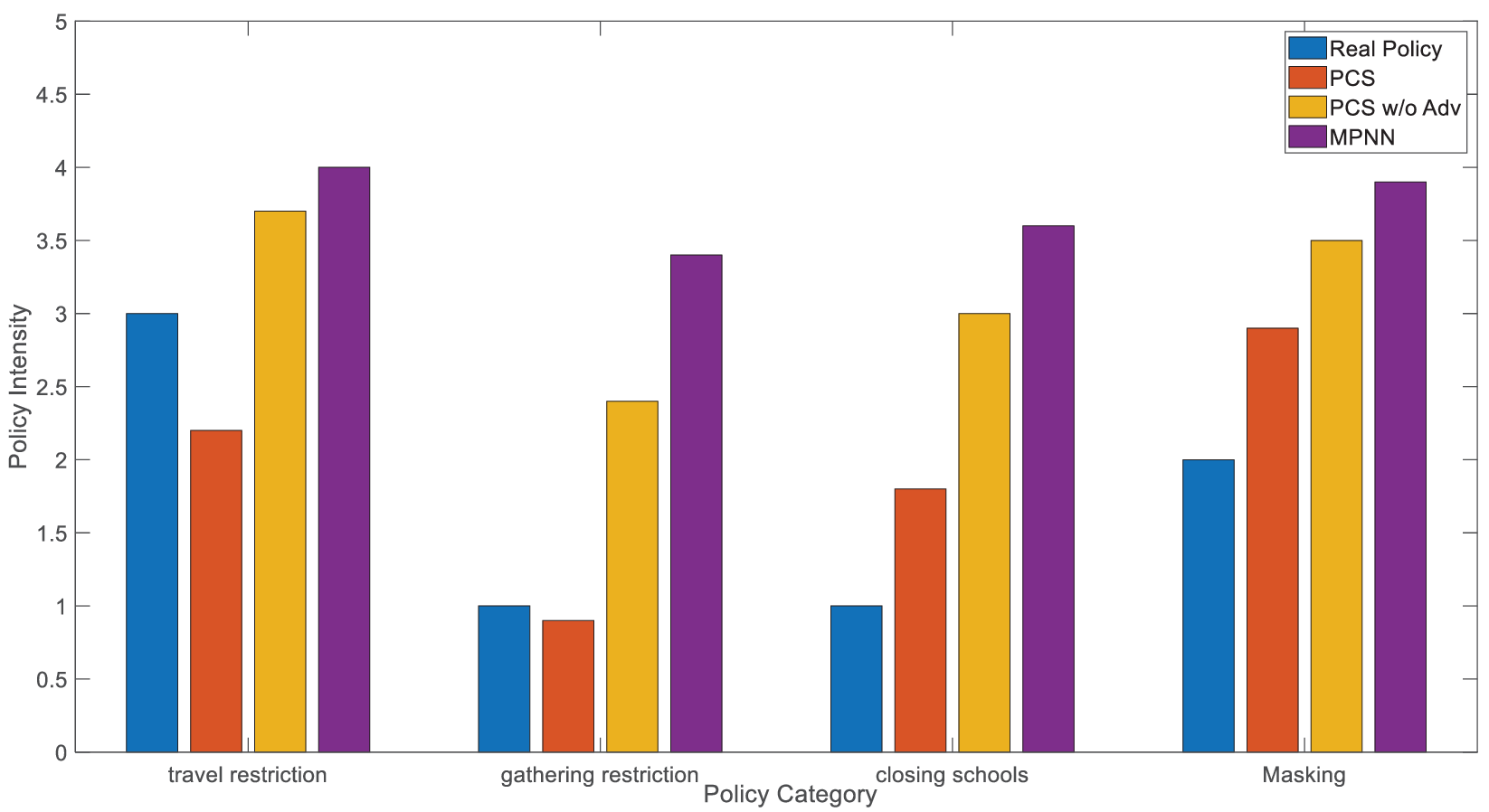}
    }
    \subfloat[Minnesota]{
    \includegraphics[width=1.1in]{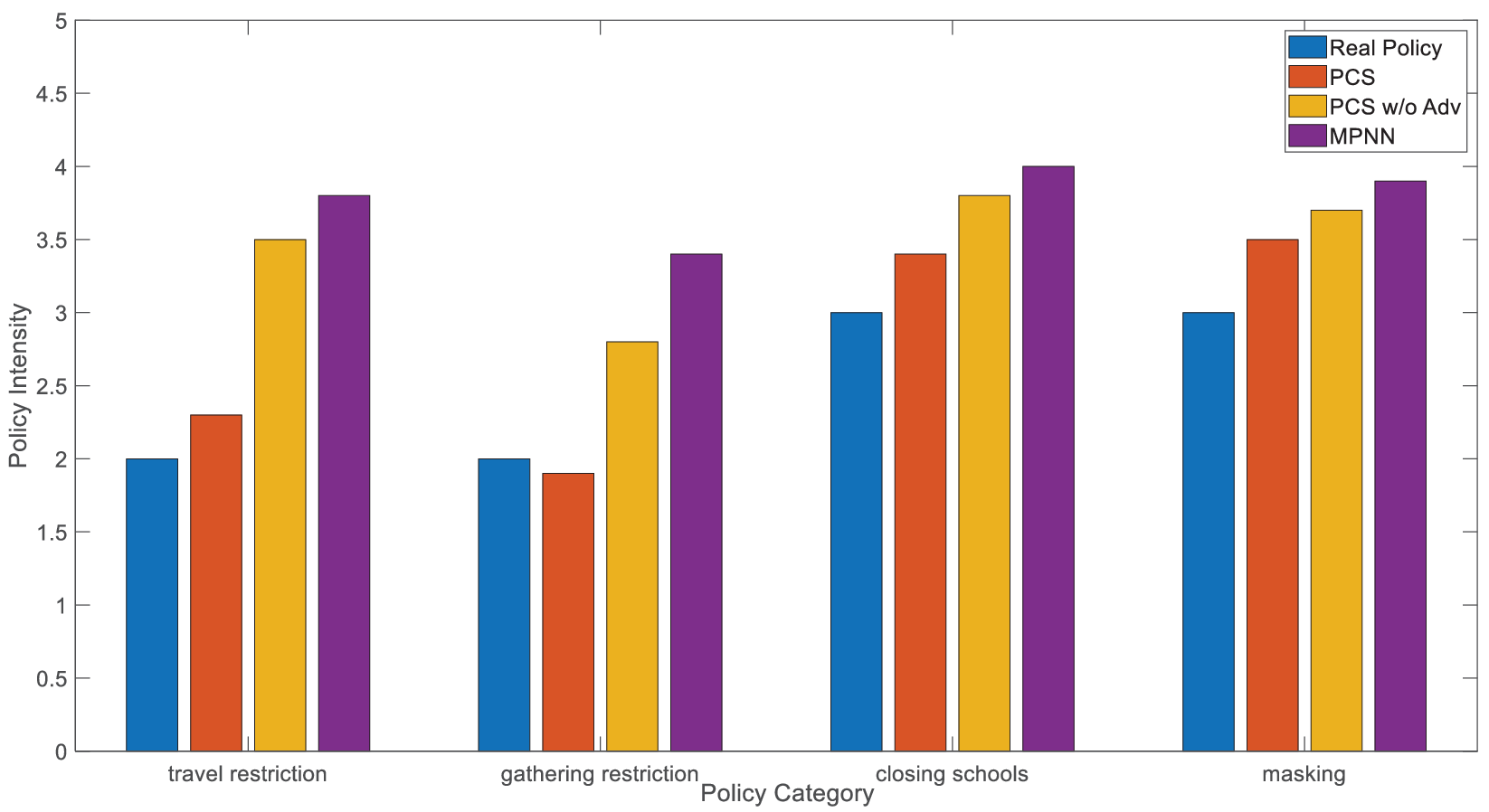}
    }
    \subfloat[Maine]{
    \includegraphics[width=1.1in]{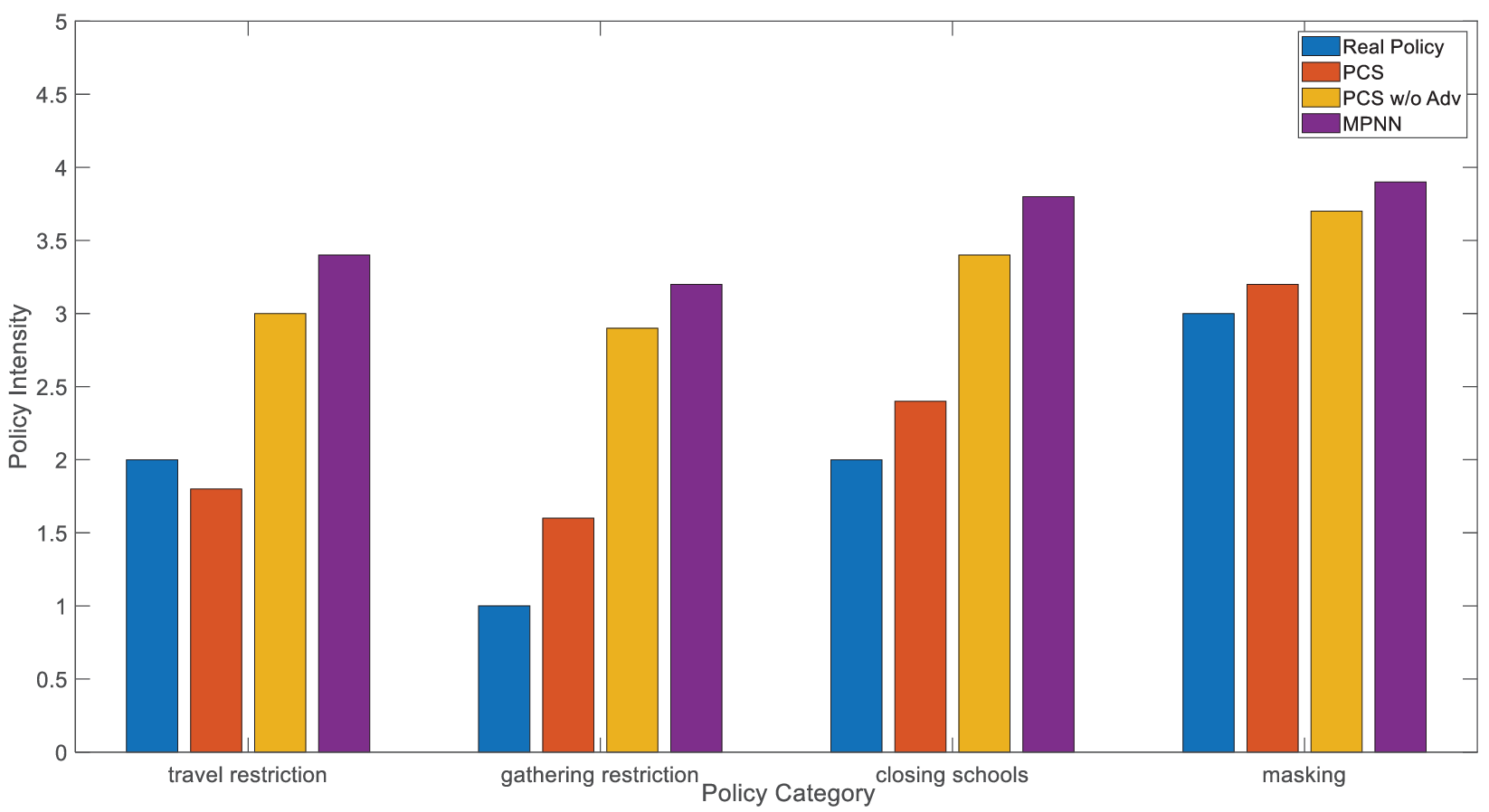}
    }
    \subfloat[Oklahoma]{
    \includegraphics[width=1.1in]{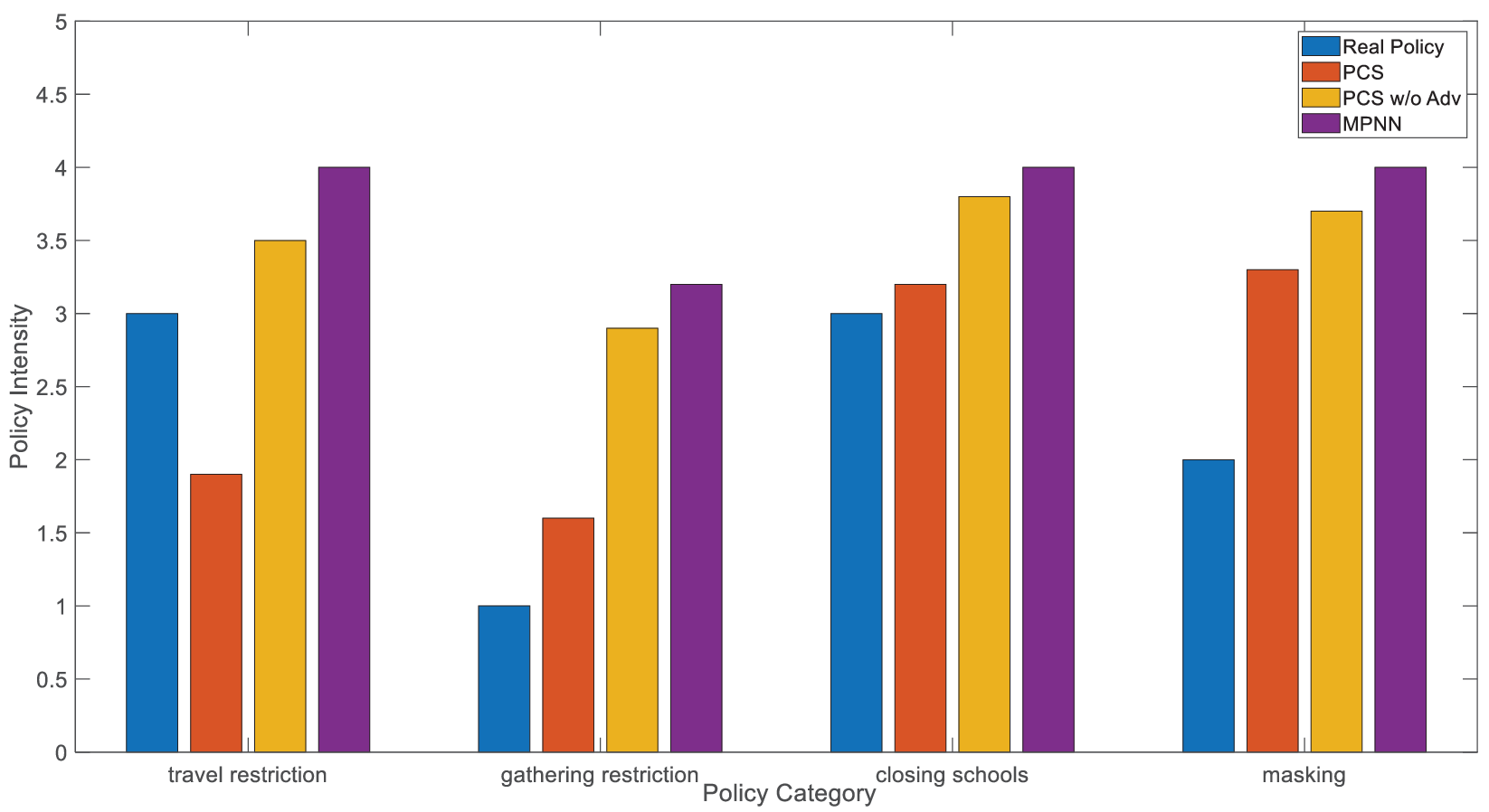}
    }
    \caption{Comparison of output policies with and without adversarial module}
    \label{fig7}
    \vspace{-20pt}
\end{figure*}

In this section, we remove the adversarial module of the PCS model and compare the output policies and their effects with those of the entire model. Meanwhile, we also add the Transformer model whose output policies are closest to the real ones among the baselines into the comparison. The results are shown in Table \ref{table4} and Fig \ref{fig7}, among which, ``PCS w/o Adv" indicates the PCS model without the adversarial module.
\begin{table*}
    \vspace{-25pt}
    \centering
    \caption{Precision of model-made policies with and without adversarial module}
    \begin{tabular}{c|c|c|c|c|c|c}\hline
        Region & Indiana & Minnesota & Maine & Oklahoma & New York & Average \\\hline
        PCS & 0.8625 & 0.8796 & 0.8667 & 0.8931 & 0.8784 & 0.8752 \\
        PCS w/o Adv & 0.8294 & 0.8335 & 0.8219 & 0.8418 & 0.8345 & 0.8302 \\
        Transformer & 0.7987 & 0.8140 & 0.8201 & 0.8011 & 0.8026 & 0.8090 \\\hline
    \end{tabular}
    \label{table4}
    \vspace{-15pt}
\end{table*}

From Table \ref{table4}, it can be found that after removing the adversarial module, policies made by the proposed model are not as close to the real ones as before. On the other hand, the proposed model's outputs without the adversarial module become closer to policies developed by baseline models (see Fig \ref{fig7}) and more radical than the real policies. These results prove that the adversarial module helps the proposed model learn the characteristics of manual decisions and prevents it from making extreme decisions.

On the other hand, we can discover that even without the adversarial module, the policies developed by the proposed model are still closer to the real ones than the best baseline models (see the comparison between PCS and Transformer in Table \ref{table4}). This is because the proposed model can still learn the critical features of human-liked decision-making through the contrast module, thereby preserving some human style in the policy-making process.

\subsection{Ablation Study on Contrast Module (RQ4)}
\begin{figure*}[btp]
    \centering
    \subfloat[Indiana]{
    \includegraphics[width=1.1in]{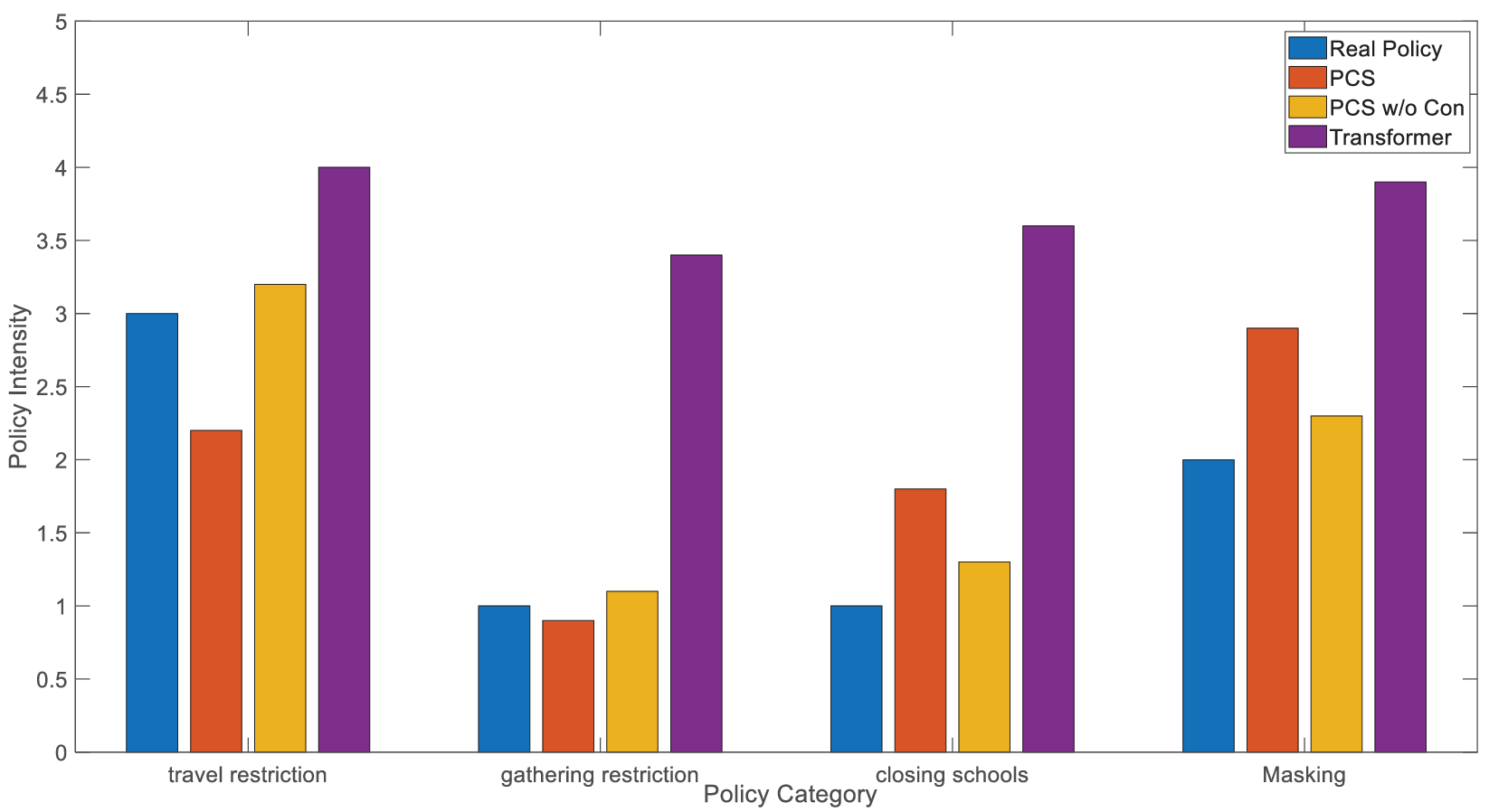}
    }
    \subfloat[Minnesota]{
    \includegraphics[width=1.1in]{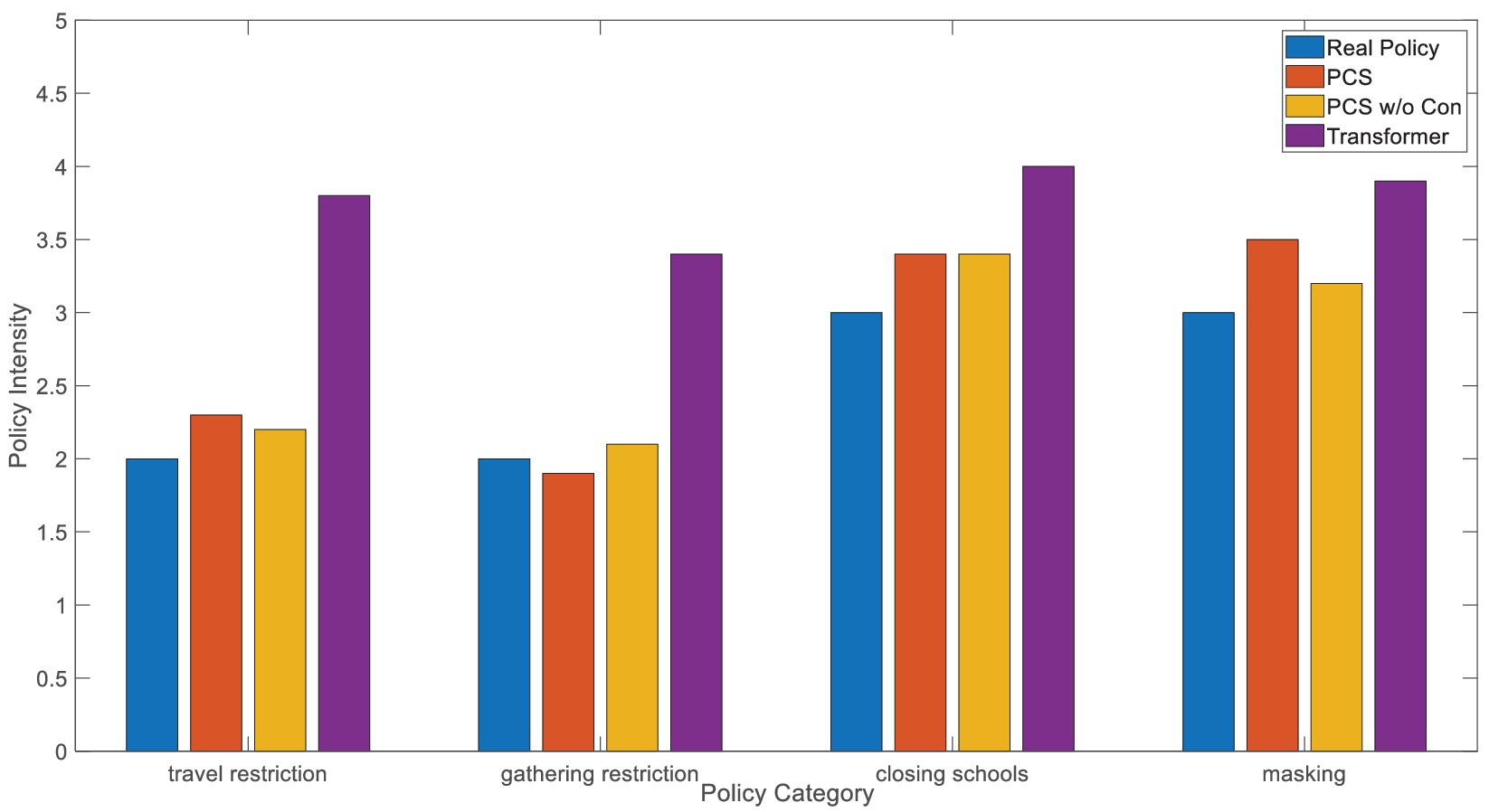}
    }
    \subfloat[Maine]{
    \includegraphics[width=1.1in]{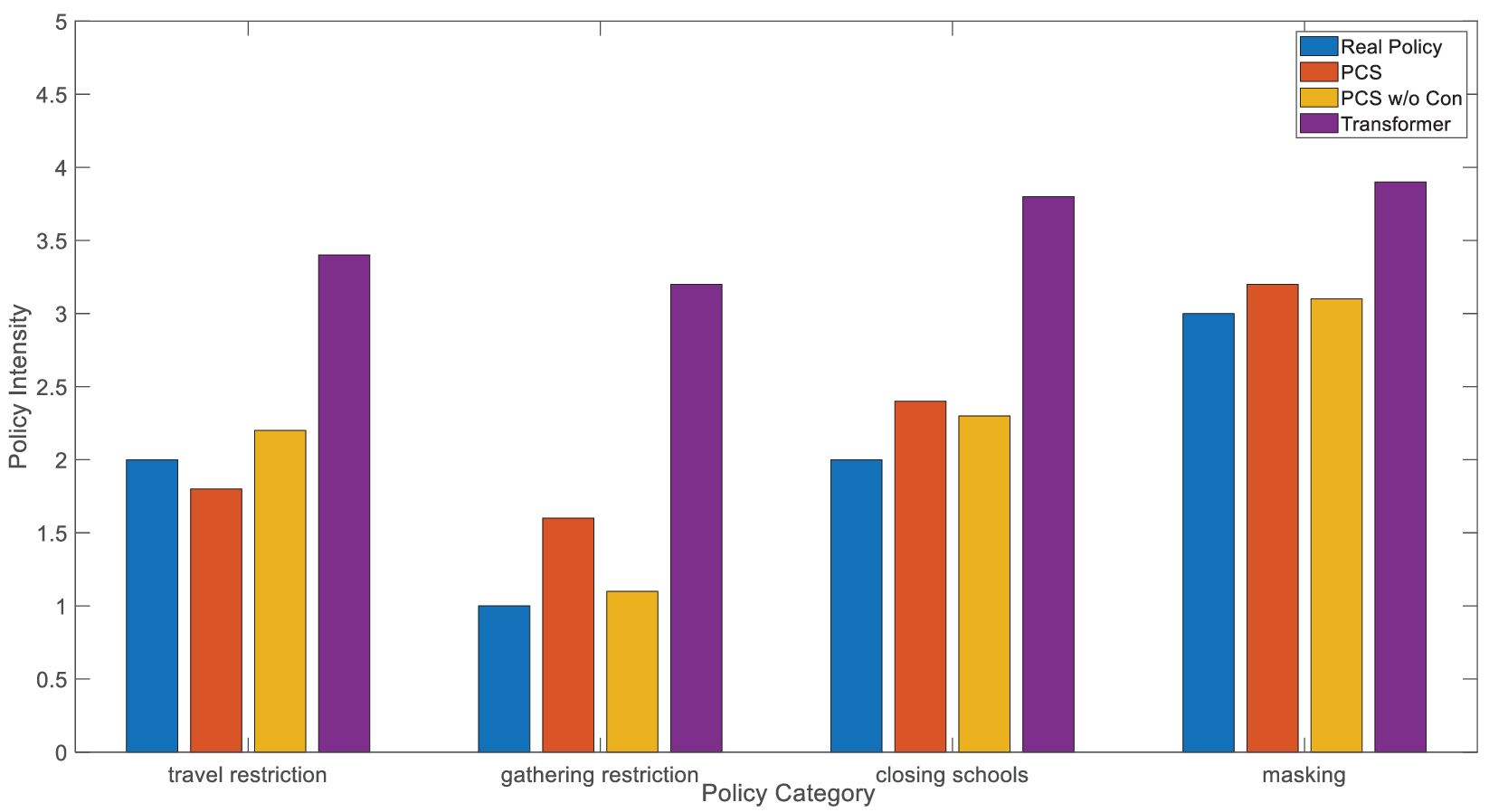}
    }
    \subfloat[Oklahoma]{
    \includegraphics[width=1.1in]{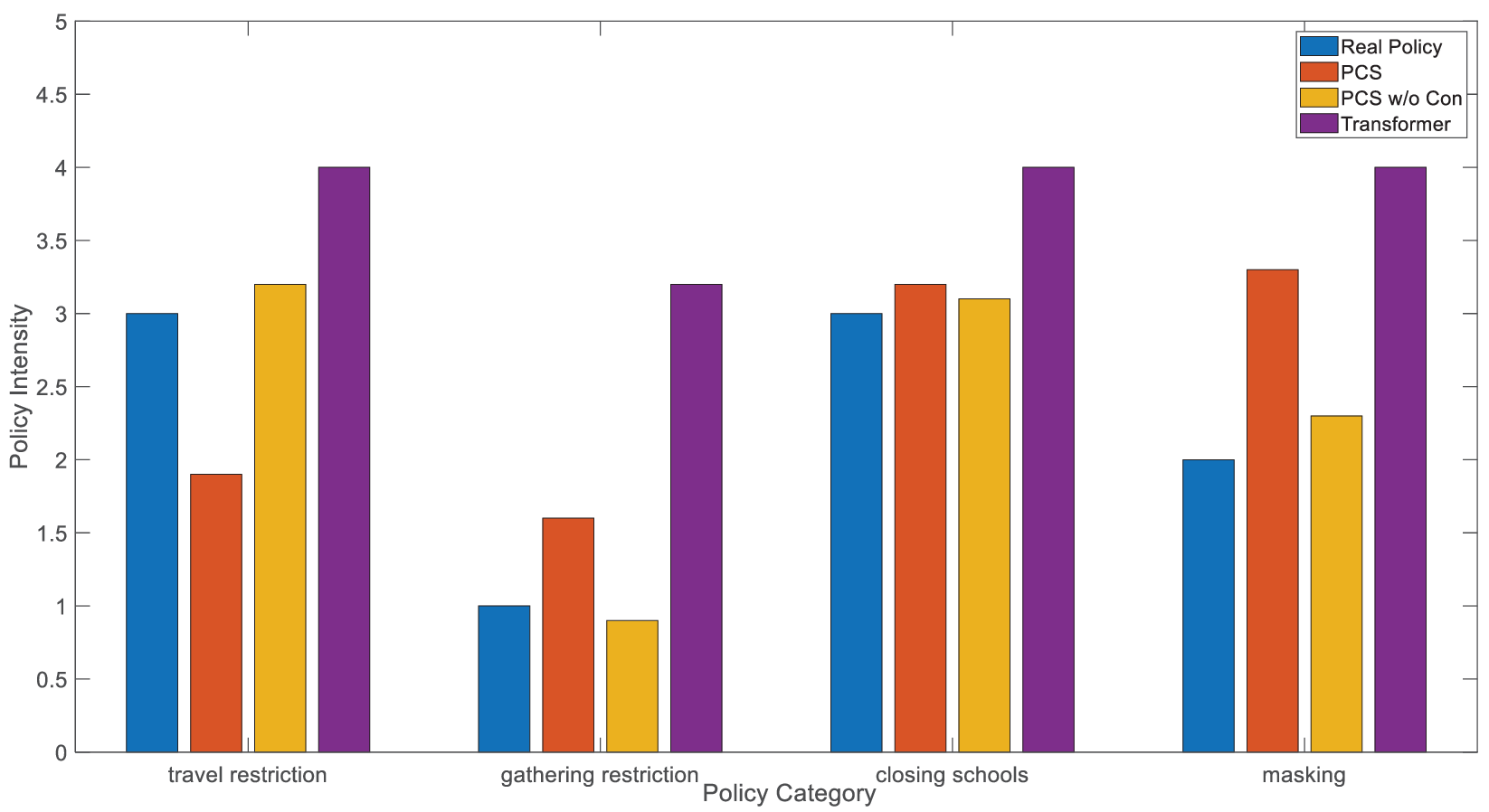}
    }
    \caption{Comparison of output policies with and without contrast module}
    \label{fig8}
    \vspace{-20pt}
\end{figure*}

The last part of the experimental results investigates the influence of the contrast module on the performance of the proposed model. To do so, we conduct similar experiments to section \ref{sec4-6} and display the results in Table \ref{table5} and Fig \ref{fig8}. (PCS w/o Con indicates PCS model with contrast module removed)
\begin{table*}
    \vspace{-25pt}
    \centering
    \caption{Precision of model-made policies with and without contrast module}
    \begin{tabular}{c|c|c|c|c|c|c}\hline
        Region & Indiana & Minnesota & Maine & Oklahoma & New York & Average \\\hline
        PCS & 0.8625 & 0.8796 & 0.8667 & 0.8931 & 0.8784 & 0.8752 \\
        PCS w/o Con & 0.9124 & 0.9201 & 0.9233 & 0.9028 & 0.8997 & 0.9190 \\
        Transformer & 0.7987 & 0.8140 & 0.8201 & 0.8011 & 0.8026 & 0.8090 \\\hline
    \end{tabular}
    \label{table5}
    \vspace{-15pt}
\end{table*}

From the results in Table \ref{table5}, it seems that without the contrast module, the proposed model can develop policies that are closer to the real ones. But if we further look into the results in Fig \ref{fig8}, it can be found that the policies made by the proposed model in Table \ref{table5} are more like the blind imitation of the real ones. In particular, when the corresponding real policies are not good or even wrong, the proposed model can hardly develop satisfied policies. This is because, without the contrast module, the proposed model loses the optimal policy-making experience under certain scenarios drawn from the global range. When the local historical policies are of limited quality, there is nothing that can help the proposed model to do better. The results in this section also illustrate why the policies developed by the proposed model can even achieve a higher comprehensive effect than the real policies in Fig \ref{fig3} and Fig \ref{fig4}.

\section{Conclusion}
In this article, we propose a novel policy-making model for epidemic prevention and control. The kernel of the proposed model is the adaptive multi-task learning framework which employs an adaptive adversarial module to learn human decision-making style and an adaptive contrast module to draw on optimal experience from global historical policies. Extensive experimental results on real-world COVID-19 data prove the effectiveness of our model.

Our model is highly scalable, every component of our model can be replaced with a more advanced method. For example, a better policy generator, more objectives for the evaluator, and more effective adversarial or contrast module design can all benefit the performance of the proposed model. Meanwhile, bringing the spreading features of other viruses into the II-SEIR model will help expand the proposed model to the policy-making task for more epidemics.
%
%
%
\bibliographystyle{splncs04}
\bibliography{mcs_ref}

\begin{thebibliography}{10}
\providecommand{\url}[1]{\texttt{#1}}
\providecommand{\urlprefix}{URL }
\providecommand{\doi}[1]{https://doi.org/#1}

\bibitem{ANNAS2020110072}
Annas, S., {Isbar Pratama}, M., Rifandi, M., Sanusi, W., Side, S.: Stability analysis and numerical simulation of seir model for pandemic covid-19 spread in indonesia. Chaos, Solitons \& Fractals  \textbf{139},  110072 (2020)

\bibitem{BOYER2022100620}
Boyer, C.B., Rumpler, E., Kissler, S.M., Lipsitch, M.: Infectious disease dynamics and restrictions on social gathering size. Epidemics  \textbf{40},  100620 (2022)

\bibitem{10.1371/journal.pone.0236310}
Canabarro, A., Tenório, E., Martins, R., Martins, L., Brito, S., Chaves, R.: Data-driven study of the covid-19 pandemic via age-structured modelling and prediction of the health system failure in brazil amid diverse intervention strategies. PLOS ONE  \textbf{15}(7),  1--13 (07 2020)

\bibitem{10.1145/3534678.3542673}
Chen, J., Hoops, S., Marathe, A., Mortveit, H., Lewis, B., Venkatramanan, S., Haddadan, A., Bhattacharya, P., Adiga, A., Vullikanti, A., Srinivasan, A., Wilson, M.L., Ehrlich, G., Fenster, M., Eubank, S., Barrett, C., Marathe, M.: Effective social network-based allocation of covid-19 vaccines. p. 4675–4683. KDD '22, Association for Computing Machinery, New York, NY, USA (2022)

\bibitem{chinazzi2020effect}
Chinazzi, M., Davis, J.T., Ajelli, M., Gioannini, C., Litvinova, M., Merler, S., Pastore~y Piontti, A., Mu, K., Rossi, L., Sun, K., et~al.: The effect of travel restrictions on the spread of the 2019 novel coronavirus (covid-19) outbreak. Science  \textbf{368}(6489),  395--400 (2020)

\bibitem{10.1145/3394486.3412863}
Ghamizi, S., Rwemalika, R., Cordy, M., Veiber, L., Bissyand\'{e}, T.F., Papadakis, M., Klein, J., Le~Traon, Y.: Data-driven simulation and optimization for covid-19 exit strategies. p. 3434–3442. KDD '20, Association for Computing Machinery, New York, NY, USA (2020)

\bibitem{10.1145/3394486.3412860}
Hao, Q., Chen, L., Xu, F., Li, Y.: Understanding the urban pandemic spreading of covid-19 with real world mobility data. p. 3485–3492. KDD '20, Association for Computing Machinery, New York, NY, USA (2020)

\bibitem{9256562}
Hassan, A., Shahin, I., Alsabek, M.B.: Covid-19 detection system using recurrent neural networks. In: 2020 International Conference on Communications, Computing, Cybersecurity, and Informatics (CCCI). pp.~1--5 (2020)

\bibitem{he2020seir}
He, S., Peng, Y., Sun, K.: Seir modeling of the covid-19 and its dynamics. Nonlinear dynamics  \textbf{101},  1667--1680 (2020)

\bibitem{HERNANDEZMATAMOROS2020106610}
Hernandez-Matamoros, A., Fujita, H., Hayashi, T., Perez-Meana, H.: Forecasting of covid19 per regions using arima models and polynomial functions. Applied Soft Computing  \textbf{96},  106610 (2020)

\bibitem{KARA2021115153}
Kara, A.: Multi-step influenza outbreak forecasting using deep lstm network and genetic algorithm. Expert Systems with Applications  \textbf{180},  115153 (2021)

\bibitem{Lin_Li_Zheng_Cheng_Yuan_2020}
Lin, Z., Li, M., Zheng, Z., Cheng, Y., Yuan, C.: Self-attention convlstm for spatiotemporal prediction. Proceedings of the AAAI Conference on Artificial Intelligence  \textbf{34}(07),  11531--11538 (Apr 2020)

\bibitem{10.1145/3485447.3512139}
Ma, J., Dong, Y., Huang, Z., Mietchen, D., Li, J.: Assessing the causal impact of covid-19 related policies on outbreak dynamics: A case study in the us. p. 2678–2686. WWW '22, Association for Computing Machinery, New York, NY, USA (2022)

\bibitem{NEURIPS2020_a6b964c0}
Min, Y., Wenkel, F., Wolf, G.: Scattering gcn: Overcoming oversmoothness in graph convolutional networks. In: Larochelle, H., Ranzato, M., Hadsell, R., Balcan, M., Lin, H. (eds.) Advances in Neural Information Processing Systems. vol.~33, pp. 14498--14508. Curran Associates, Inc. (2020)

\bibitem{info11090454}
Pirouz, B., Nejad, H.J., Violini, G., Pirouz, B.: The role of artificial intelligence, mlr and statistical analysis in investigations about the correlation of swab tests and stress on health care systems by covid-19. Information  \textbf{11}(9) (2020)

\bibitem{NEURIPS2020_79a3308b}
Qian, Z., Alaa, A.M., van~der Schaar, M.: When and how to lift the lockdown? global covid-19 scenario analysis and policy assessment using compartmental gaussian processes. In: Larochelle, H., Ranzato, M., Hadsell, R., Balcan, M., Lin, H. (eds.) Advances in Neural Information Processing Systems. vol.~33, pp. 10729--10740. Curran Associates, Inc. (2020)

\bibitem{9705122}
Tayarani-Najaran, M.H.: A novel ensemble machine learning and an evolutionary algorithm in modeling the covid-19 epidemic and optimizing government policies. IEEE Transactions on Systems, Man, and Cybernetics: Systems  \textbf{52}(10),  6362--6372 (2022)

\bibitem{9613774}
Tutsoy, O.: Pharmacological, non-pharmacological policies and mutation: An artificial intelligence based multi-dimensional policy making algorithm for controlling the casualties of the pandemic diseases. IEEE Transactions on Pattern Analysis and Machine Intelligence  \textbf{44}(12),  9477--9488 (2022)

\bibitem{NIPS2017_3f5ee243}
Vaswani, A., Shazeer, N., Parmar, N., Uszkoreit, J., Jones, L., Gomez, A.N., Kaiser, L.u., Polosukhin, I.: Attention is all you need. In: Guyon, I., Luxburg, U.V., Bengio, S., Wallach, H., Fergus, R., Vishwanathan, S., Garnett, R. (eds.) Advances in Neural Information Processing Systems. vol.~30. Curran Associates, Inc. (2017)

\bibitem{10001858}
Wakugawa, M., Saitoh, F.: Impact of covid-19 asymptomatic individuals on effective regenerative math by multi-agent simulation based on the seair model. In: 2022 Joint 12th International Conference on Soft Computing and Intelligent Systems and 23rd International Symposium on Advanced Intelligent Systems (SCIS\&ISIS). pp.~1--4 (2022)

\bibitem{zhang2018overview}
Zhang, Y., Yang, Q.: An overview of multi-task learning. National Science Review  \textbf{5}(1),  30--43 (2018)

\end{thebibliography}


\begin{thebibliography}{8}
\bibitem{ref_article1}
Author, F.: Article title. Journal \textbf{2}(5), 99--110 (2016)

\bibitem{ref_lncs1}
Author, F., Author, S.: Title of a proceedings paper. In: Editor,
F., Editor, S. (eds.) CONFERENCE 2016, LNCS, vol. 9999, pp. 1--13.
Springer, Heidelberg (2016). \doi{10.10007/1234567890}

\bibitem{ref_book1}
Author, F., Author, S., Author, T.: Book title. 2nd edn. Publisher,
Location (1999)

\bibitem{ref_proc1}
Author, A.-B.: Contribution title. In: 9th International Proceedings
on Proceedings, pp. 1--2. Publisher, Location (2010)

\bibitem{ref_url1}
LNCS Homepage, \url{http://www.springer.com/lncs}, last accessed 2023/10/25
\end{thebibliography}
%
\end{document}